\documentclass[journal=jacsat,manuscript=article]{achemso}
\SectionNumbersOn

\usepackage[version=3]{mhchem} 
\usepackage{graphicx}
\usepackage{amsfonts}
\usepackage{xcolor}
\usepackage{hyperref}       
\usepackage{url}            
\usepackage{booktabs}       
\usepackage{microtype}      
\usepackage{amsmath}
\usepackage{subcaption}
\usepackage{enumitem}
\usepackage{multirow}
\usepackage{tikz}
\usepackage{makecell}

\newcommand{\angstrom}{\mbox{\normalfont\AA}}

\author{Yuyang Wang}
\affiliation{Department of Mechanical Engineering, Carnegie Mellon University, Pittsburgh, PA 15213, USA}
\alsoaffiliation{Machine Learning Department, Carnegie Mellon University, Pittsburgh, PA 15213, USA}
\author{Changwen Xu}
\affiliation{Department of Mechanical Engineering, Carnegie Mellon University, Pittsburgh, PA 15213, USA}
\author{Zijie Li}
\affiliation{Department of Mechanical Engineering, Carnegie Mellon University, Pittsburgh, PA 15213, USA}
\author{Amir Barati Farimani}
\affiliation{Department of Mechanical Engineering, Carnegie Mellon University, Pittsburgh, PA 15213, USA}
\alsoaffiliation{Machine Learning Department, Carnegie Mellon University, Pittsburgh, PA 15213, USA}
\alsoaffiliation{Department of Materials Science and Engineering, Carnegie Mellon University, Pittsburgh, PA 15213, USA}
\alsoaffiliation{Department of Chemical Engineering, Carnegie Mellon University, Pittsburgh, PA 15213, USA}
\email{barati@cmu.edu}

\title[An \textsf{achemso} demo]{Denoise Pretraining on Nonequilibrium Molecules for Accurate and Transferable Neural Potentials}

\abbreviations{IR,NMR,UV}

\begin{document}




\begin{abstract}

Recent advances in equivariant graph neural networks (GNNs) have made deep learning amenable to developing fast surrogate models to expensive \textit{ab initio} quantum mechanics (QM) approaches for molecular potential predictions. However, building accurate and transferable potential models using GNNs remains challenging, as the data is greatly limited by the expensive computational costs and level of theory of QM methods, especially for large and complex molecular systems. In this work, we propose denoise pretraining on nonequilibrium molecular conformations to achieve more accurate and transferable GNN potential predictions. Specifically, atomic coordinates of sampled nonequilibrium conformations are perturbed by random noises and GNNs are pretrained to denoise the perturbed molecular conformations which recovers the original coordinates. Rigorous experiments on multiple benchmarks reveal that pretraining significantly improves the accuracy of neural potentials. Furthermore, we show that the proposed pretraining approach is model-agnostic, as it improves the performance of different invariant and equivariant GNNs. Notably, our models pretrained on small molecules demonstrate remarkable transferability, improving performance when fine-tuned on diverse molecular systems, including different elements, charged molecules, biomolecules, and larger systems. These results highlight the potential for leveraging denoise pretraining approaches to build more generalizable neural potentials for complex molecular systems.

\end{abstract}

\section{Introduction}

The development of efficient and transferable molecular potentials plays a key role in performing accurate simulations of molecular systems. However, accurate and transferable \textit{ab initio} quantum mechanical (QM) methods, like Hartree-Fock methods \cite{monkhorst1977calculation, hirata2004coupled, cremer2011moller} and density functional theory (DFT) \cite{levy1979universal, thanthiriwatte2011assessment}, are computationally expensive which limits simulations for large and complex systems. On the other hand, prevalent force fields, built on fitting data from QM calculations or experiments, are computationally efficient while usually suitable for only specific systems. Efforts have been made to develop empirical force fields that work for certain applications \cite{mark2001structure}, including small organic molecules \cite{halgren1996merck}, biomolecules \cite{salomon2013overview, huang2013charmm36}, and materials \cite{sun1998compass}. Nevertheless, these curated methods may fail to model complex systems involving significant polarization and many-body interactions. Also, they can be difficult to transfer to different systems, as empirical force fields are tailored to match the data from QM or experiments of different systems \cite{vitalini2015dynamic, harrison2018review, unke2020high}. How to design accurate and transferable approximations of \textit{ab initio} QM methods has been one of the major challenges in modern computational chemistry. 

Machine learning (ML) has emerged as a powerful tool to learn interatomic potentials or force fields by fitting the data in computational chemistry, promising near-QM accuracy with faster computations \cite{butler2018machine, unke2021machine}. Physical constraints, including energy conservation and roto-translational invariance, are required for ML potentials \cite{li2022graph, fu2022forces, atz2021geometric}, and various approaches have been developed to meet these constraints. For example, Gradient domain ML (GDML) predicts force fields directly and applies a kernel-based method to guarantee conservative energies \cite{chmiela2017machine}. Symmetric GDML (sGDML) extends GDML to incorporate space group symmetries and dynamic nonrigid symmetries \cite{chmiela2018towards}. Gaussian approximation potentials (GAPs) learn an energy decomposition as a summation of each atomic-centered environment \cite{bartok2010gaussian, bartok2017machine}, and different kernels or descriptors for local atomic contributions are also investigated \cite{bartok2013representing, grisafi2018symmetry}. 

Neural potentials use deep neural networks (DNNs) to model the relationship between molecular geometries and potential energy surface (PES) \cite{fedik2022extending}. DNNs are built on top of geometric descriptors of molecular systems to model potentials. Initially, interatomic distances and angles were used, but such descriptors fail to express the permutational invariance of atomic order in the input \cite{blank1995neural, brown1996combining}. High-dimensional neural network potential (HDNNP) \cite{behler2007generalized} was introduced, which breaks down the molecular system into local atomic environments. These atomic environments are encoded using atom-centered symmetry functions (ACSFs) \cite{behler2011atom} consisting of two- and three-body terms. The HDNNP has been extended by ANI \cite{smith2017ani, smith2018less, devereux2020extending} and TensorMol \cite{yao2018tensormol}. However, such models are limited in transferability, namely, reparameterization for new chemical elements requires retraining the whole model. It is notable that the size of the descriptor grows quadratically with the number of elements. 

End-to-end neural potentials, on the other hand, are constructed directly on Cartesian coordinates and element types, rather than on handcrafted descriptors. Graph neural networks (GNNs) based on message-passing \cite{gilmer2017neural} are utilized to learn representations from molecular structures, with nodes representing atoms and edges representing interatomic interactions. MLPs are implemented on top of the representations learned by GNNs to model the potentials. GNNs learn chemical and physical interactions through aggregating neighboring atomic information and updating the nodes. SchNet implements continuous convolution filters to model interatomic distances \cite{schutt2018schnet} and DimeNet incorporates angles to model three-body interactions in message-passing \cite{Gasteiger2020Directional}. Other works attempt to include chemical domain knowledge to develop better neural potentials \cite{unke2019physnet, lubbers2018hierarchical, zubatyuk2019accurate}. However, the message-passing operations in these models fail to encode directional information. In neural potentials, we are interested in the equivariance with respect to three-dimensional (3D) rigid-body transformation, such that the neural network commutes with any 3D rotations, translations, and/or reflections. The first category of equivariant GNNs is based on irreducible representations in group theory including Tensor field network (TFN) \cite{thomas2018tensor}, Cormorant \cite{anderson2019cormorant}, SE(3)-Transformer \cite{fuchs2020se}, SEGNN \cite{brandstetter2022geometric}, etc. These models build equivariant via Clebsch Gordon coefficients and spherical harmonic basis as well as learnable radial neural networks. The second category of equivariant GNNs relies on linear operations (i.e., scaling, linear combination, dot product, and vector product) on vectorial features instead \cite{jing2021learning, villar2021scalars, gasteiger2021gemnet, batzner20223}. PaiNN \cite{schutt2021equivariant} and TorchMD-Net \cite{tholke2022torchmd} are two examples of such equivariant GNNs. PaiNN proposes to keep track of scale features and vector features separately and develops linear operations to pass information in between to keep the equivariance. TorchMD-Net extends such an architecture with the multi-head self-attention mechanism \cite{vaswani2017attention}. In our work, we focus on equivariant GNNs due to their superior performance in neural potential benchmarks \cite{tholke2022torchmd}.

To improve the performance of GNNs on molecular property predictions, self-supervised learning (SSL) approaches \cite{hadsell2006dimensionality, chen2020simple} have been investigated. GNNs are first pretrained via SSL to learn expressive molecular representations and then fine-tuned on downstream prediction tasks. Predictive SSL methods rely on recovering the original instance from partially observed or perturbed samples, including masked attribute prediction \cite{Hu2020Strategies}, context prediction \cite{rong2020self}, motif tree generation \cite{zhang2021motif}. Also, GEM \cite{fang2022geometry} proposes to pretrain GNNs via predicting 3D positional information, including interatomic distances, bond lengths, and bond angles. On the other hand, contrastive learning, which aims at learning representations via contrasting positive instances against negative instances, has been widely implemented in GNNs for chemical sciences \cite{liu2022graph, krishnan2022self, magar2022crystal, cao2022moformer}. MolCLR \cite{wang2022molecular} applies random masking of atom and edge attributes to generate contrastive instances of molecular graphs. Recent works have also investigated multi-level subgraphs in contrastive training \cite{zhang2020motif, wang2022improving}. GraphMVP \cite{liu2022pretraining} and 3D infomax \cite{stark20223d} incorporate 3D information into 2D graphs via contrasting molecules represented as 2D topological graphs and 3D geometric structures. It is noted that most SSL works neglect 3D information. Even though a few SSL approaches incorporate 3D information, they are not built upon equivariant GNNs, which greatly limits the applications to accurate neural potential predictions. Recently, a few works propose denoising as an SSL approach to pretrain GNNs \cite{zaidi2022pre, liu2022molecular, zhou2023unimol}. By predicting the noise added to atomic coordinates at the equilibrium states, GNNs are trained to learn a particular pseudo-force field. Such a method has demonstrated effectiveness on QM property prediction benchmarks like QM9 \cite{ruddigkeit2012enumeration}. Nevertheless, it only leverages molecules at equilibrium states which is far from sufficient for accurate neural potentials, since potential predictions require evaluation of nonequilibrium molecular structures in simulations. Overall, SSL has not been well studied for neural potential predictions. 

\begin{figure}[t!]
    \centering
    \includegraphics[width=\textwidth, keepaspectratio=true]{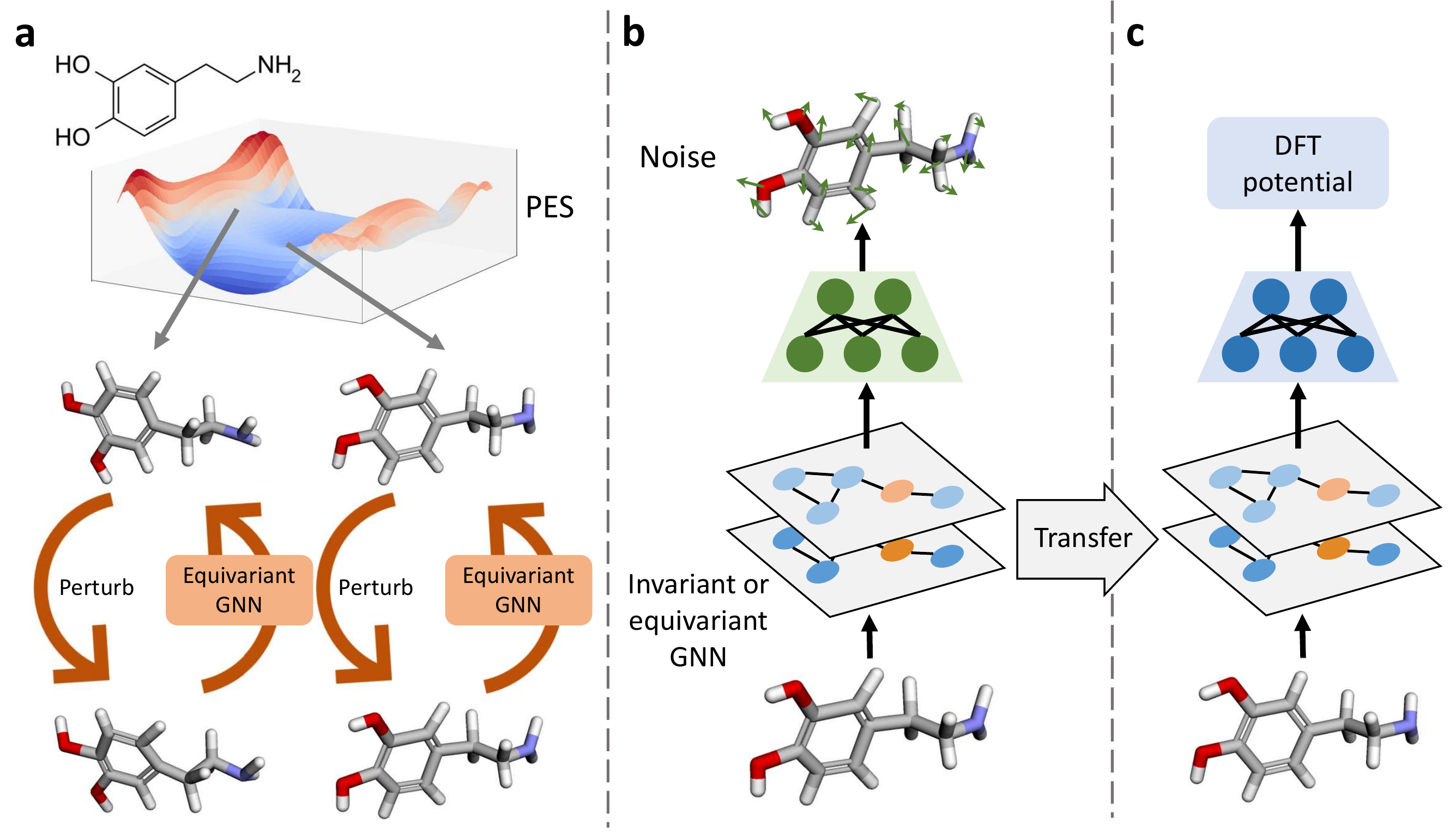}
    \caption{Framework of pretraining invariant and equivariant GNNs on nonequilibrium molecules for potential predictions. (a) Different molecular conformations are sampled and random noises are added to perturb the conformations. GNN is trained to recover the original conformations from perturbed ones. (b) During pretraining, perturbed conformations are fed into the GNN to predict the additional noise. (3) The pretrained GNN is transferred and fine-tuned with an MLP head to predict potential energies calculated by DFT. }
    \label{fig:framework}
\end{figure}

Although neural potentials have been extensively investigated, their accuracy and transferability are still limited due to the reliance on expensive QM calculations to obtain training data. In particular, for large and complex molecular systems, collecting accurate and sufficient QM data can be extremely challenging or even infeasible. This raises two questions: 1) can we use SSL to improve the accuracy of neural potentials with currently available data; and 2) can we use SSL to develop more transferable neural potentials that leverage relatively rich molecular potential data (e.g., small molecules) and apply pretrained models to large and complex molecular systems with limited data? To address these challenges, we propose an SSL pretraining strategy on nonequilibrium molecules to improve the accuracy and transferability of neural potentials (Figure~\ref{fig:framework}(a)). This involves adding random noises to the atomic coordinates of molecular systems at nonequilibrium states, and training GNNs to predict the artificial noises in an SSL manner (Figure~\ref{fig:framework}(b)). We then fine-tune the pretrained models on multiple challenging molecular potential prediction benchmarks (Figure~\ref{fig:framework}(c)). Our rigorous experiments demonstrate that our proposed pretraining method, which leverages nonequilibrium molecules, significantly improves the accuracy of neural potentials and is model-agnostic to various invariant or equivariant GNN architectures, including SchNet, SE(3)-Transformer, EGNN, and TorchMD-Net. Notably, GNNs pretrained on small molecules demonstrate significant improvement in potential predictions when fine-tuned on diverse molecular systems, including those with different elements, charged molecules, biomolecules, and larger systems. Such a denoise pretraining approach has the potential to facilitate the development of accurate and transferable surrogate models for expensive QM methods, and enable the application of neural potentials to complex systems.


\section{Method}

\subsection{Graph Neural Networks}

A molecular system can be modeled as a graph $\mathcal{G} = (V, \mathcal{A})$, where V denotes the $N$ atoms in the system and $\mathcal{A}$ denotes the interactions between atoms (e.g., bonds). To learn the representation of the molecular graph, modern GNNs utilize message-passing strategies \cite{gilmer2017neural}. Let ${h}_i^{(t)} \in \mathbb{R}^{F}$ be the F-dimensional feature of atom $i$ at the $t$-th layer of GNN and ${a}_{ij}$ represent the interaction between atoms $i$ and $j$. For atom $i$, the message ${m}_{ij}^{(t)}$ passed from its neighboring atoms $j \in \mathcal{N}_i$ is aggregated via the function $f_m^{(t)}$ (Eq.~\ref{eq:message}). All the $m_{ij}$ are summed as the message passed to node $i$ , followed by an update function $f_u^{(t)}$ that updates the atomic feature with the aggregated message from neighbors and its original feature (Eq.~\ref{eq:update}).  
\begin{equation}
    {m}_{ij}^{(t)} = f_m^{(t)}(h_i^{(t)}, h_j^{(t)}, a_{ij}),
    \label{eq:message}
\end{equation}
\begin{equation}
     h_i^{(t+1)} = f_u^{(t)} (h_i^{(t)}, \sum_{j \in \mathcal{N}_i} m_{ij}^{(t)}).
    \label{eq:update}
\end{equation}

\subsection{Equivariant Graph Neural Networks}

In this work, we investigate GNNs for molecular potential predictions, which require 3D positional information of atoms. To this end, equivariant GNNs are introduced which extend the message-passing functions to keep the physical symmetry of the molecular systems. Let $T_g: \mathcal{X} \rightarrow \mathcal{X}$ define a set of transformations of the abstract group $g \in G$ in input space $X$ and $\phi: \mathcal{X} \rightarrow \mathcal{Y}$ is a function that maps the input to the output space $\mathcal{Y}$. A function $\phi$ is equivariant to $g$ if there exists a transformation $S_g: \mathcal{Y} \rightarrow \mathcal{Y}$ such that $\phi(T_g(x)) = S_g(\phi(x))$, $\forall g \in G$ and $x \in \mathcal{X}$. E(n)-equivariance denotes equivariance with Euclidean group E(n) which comprises all translations, rotations, and reflections in n-dimensional Euclidean space, while SE(n)-equivariance only satisfies translation and rotation equivariance \cite{satorras2021n}. Equivariant GNNs introduce the equivariance as an inductive bias to molecular modeling and demonstrate superior performance in many energy-related property predictions. 

One strategy of designing equivariant message-passing operations is based on irreducible representations in group theory \cite{anderson2019cormorant, fuchs2020se}. TFN \cite{thomas2018tensor} introduces the Clebsch-Gordon coefficient, radial neural network, and spherical harmonics as building blocks for SE(3)-equivariant message passing as given in Eq.~\ref{eq:tfn} and \ref{eq:clebsch} (we ignore the superscript of the layer to simplify the notation).
\begin{equation}
    {m}_{ij} = \sum_{k \geq 0} {W}^{\ell k} ({r}_{ji}) {h}_j^{k},
\label{eq:tfn}
\end{equation}
\begin{equation}
    {W}^{\ell k}({x}) = \sum_{J=|k-\ell|}^{k+\ell} \varphi_J^{\ell k}(\|{x}\|) \sum_{m=-J}^J Y_{Jm} ({x} / \|{x}\|) {Q}_{Jm}^{\ell k},
\label{eq:clebsch}
\end{equation}
where ${r}_{ij}={x}_i - {x}_j$ is the directional vector between the Cartesian coordinates ${x}_i$ and ${x}_j$ of two atoms $i$ and $j$, ${h}_j^{k}$ is a type-$k$ feature of node $j$, and ${W}^{\ell k}: \mathbb{R}^3 \rightarrow \mathbb{R}^{(2\ell+1)\times(2k+1)}$ is the weight kernel that maps type-$k$ features to type-$\ell$ features. ${W}^{\ell k}$ is decomposed as the linear combination of non-learnable Clebsch-Gordan matrices ${Q}_{Jm}^{\ell k}$ and sphereical harmonics $Y_J: \mathbb{R}^3 \rightarrow \mathbb{R}^{2J+1}$ with a learnable radial network $\varphi$ as given in Eq.\ref{eq:clebsch}. Due to the implementation of the Clebsch-Gordon tensor product and spherical harmonics, such SE(3)-equivariant GNNs are usually computationally expensive and have a limited number of learnable parameters \cite{atz2021geometric}. 

Another method to build the equivariance is to apply only linear operations (i.e., scaling, linear combination, dot product, and vector product) to vectorial features in message-passing \cite{villar2021scalars, schutt2021equivariant}. Following the insight, a straightforward way to build E(3)-equivariant operations is to keep track of a vectorial feature $v_i \in \mathbb{R}^{3 \times F}$ besides the scalar feature $h_i$ for each node. Thus, the message passing function and update function for vectorial features are shown in Eq.~\ref{eq:egnn_message} and \ref{eq:egnn_update}, respectively.
\begin{equation}
    {m}_{ij}^{(t)} = f_m^{(t)}( h_i^{(t)},  h_j^{(t)}, v_i^{(t)}, v_j^{(t)}, \| {r}_{ij} \|, a_{ij}),
\label{eq:egnn_message}
\end{equation}
\begin{equation}
    {v}_{i}^{(t+1)} = {v}_i^{(t)} + \sum_{j \in \mathcal{N}_i} {r}_{ij} f_v^{(t)}({m}_{ij}^{(t)}),
\label{eq:egnn_update}
\end{equation}
where $f_v$ maps the message $m_{ij}$ to a scalar and the vectorial feature of each atom is updated as a linear combination of directional vectors $r_{ij}$ in each layer $t$. It should be noted that only linear operations can be applied to vectorial features in the message function $f_m$ to keep the equivariance. Usually, atoms within a cutoff distance $d_{\text{cut}}$ from atom $i$ are included in the neighboring list $\mathcal{N}_i$. Such a strategy paves a more flexible way to equivariant GNNs compared with irreducible representations. 

After $T$ equivariant message-passing layers, global pooling over all node features can be applied to obtain the representation of the whole molecule. In our work, we apply summation over all the node features as the pooling \cite{xu2018how}. The pooled representation is then fed into an MLP to predict the molecular potential as $\hat{E} = \texttt{MLP}(\sum_{i} h^{(T)}_i)$.

Besides, invariance indicates that the transformations (e.g., rotations and translations) of the input will not change the output, such that $\phi(T_g(x)) = \phi(x)$, $\forall g \in G$ and $x \in \mathcal{X}$. Invariance is a special equivariance where $S_g$ is the identity mapping for $\forall g \in G$. Building an invariant GNN is more straightforward than an equivariant one. By simply replacing Eq.~\ref{eq:message} with Eq.~\ref{eq:egnn_message}, one can obtain an E(3)-invariant GNN, since the geometric information is only embedded as the distances between atoms in message passing. 

In this study, we implement one invariant GNN (i.e., SchNet \cite{schutt2018schnet}) and three equivariant GNN models (i.e., SE(3)-Transformer \cite{fuchs2020se}, EGNN \cite{satorras2021n}, and TorchMD-Net \cite{tholke2022torchmd}) to validate the generalizability of the proposed pretraining method. SchNet adapts a continuous-filter convolution that contains element-wise multiplication between node features and a weight matrix that depends only on interatomic distances. SE(3)-Transformer falls into the first category of equivariant GNNs which extends TFN by introducing a scalar self-attention \cite{vaswani2017attention} term $\alpha_{ij}$ to Eq.~\ref{eq:tfn}. The input feature is mapped to key and query following the learnable weight kernel in Eq.~\ref{eq:clebsch} to retain the SE(3)-equivariance, and $\alpha_{ij}$ is calculated as the normalized dot product of key and query. EGNN and TorchMD-Net, on the other hand, follow the second strategy of equivariant GNNs that apply linear operations to vectorial features. EGNN directly manipulates the coordinates of atoms as vectorial features. It implements $f_m$ in Eq.~\ref{eq:egnn_message} as the concatenation of node features, edge features, and distances followed by an MLP to compute the message. The atomic coordinates are then updated by the linear combination of interatomic directional vectors weighted by the messages in each layer. TorchMD-Net explicitly models vectorial features asides from node features and develops self-attention of scalar features and interatomic distances in $f_m$. The self-attention term is utilized as a scalar term for directorial vectors between atoms to update vector features and scalar features are updated with the dot product of vectorial features. These models represent a wide variety of invariant/equivariant GNNs and have achieved competitive performance on multiple molecular benchmarks related to force fields or energies \cite{schutt2021equivariant, tholke2022torchmd}. 


\subsection{Denoising on Nonequilibrium Molecules}
\label{sec:denoise}

The pretraining strategy is based on predicting the artificial noise added to sampled conformations of molecules. A conformation of a molecule is denoted as ${V}$ and ${X}$, where ${V}$ encodes the atomic information and ${X} \in \mathbb{R}^{N \times 3}$ containing Cartesian coordinates of all $N$ atoms in the molecule. The interatomic interactions $\mathcal{A}$ can be directly derived from ${X}$ for GNN models. Random noises ${\mathcal{E}} \in \mathbb{R}^{N \times 3}$ sampled from Gaussian $\mathcal{N}({0}, \sigma I)$ are added to the position of all atoms to perturb the molecular conformation. The perturbed conformation is denoted as $\hat{{X}} = {X} + {\mathcal{E}}$. During pretraining, a GNN model is trained to predict the additional noise, and the objective function is shown in Eq.~\ref{eq:loss}.
\begin{equation}
    \mathbb{E}_{p(\hat{X}; V)} \left[ \| \phi_\theta({V}, \hat{ {X}}) - {\mathcal{E}} \|_2^2 \right],
    \label{eq:loss}
\end{equation}
where $\phi_\theta$ denotes an invariant/equivariant GNN parameterized by $\theta$ and $p(\hat{X}; V)$ measures the probability distribution of perturbed molecular conformations given the atoms. 

Such a denoising strategy in a self-supervised manner is related to learning a pseudo-force field at the perturbed states \cite{zaidi2022pre, xie2022crystal, arts2023two}. In this work, we extend such a concept from equilibrium molecular conformations to nonequilibrium ones, which is pivotal in molecular simulations. For a given molecule with $N$ atoms encoded as ${V}$, the probability of a molecular conformation ${X}$ is $p({X}; {V}) \sim \exp{(-E({X}; {V}))}$ following the Boltzmann distribution, where $E({X}; {V})$ is the potential energy. The force field of each atom in the conformation is $-\nabla_{{X}} E({X; {V}}) = -\nabla_{X} \log p( X; {V})$. The molecular conformation $ X$ can be sampled via molecular dynamic simulation, normal mode sampling, torsional sampling, etc. These methods provide diverse physically and chemically feasible nonequilibrium conformations around the equilibrium states which help understand the energetics of molecular systems \cite{smith2017ani}. It should be noted that in our case, we refer to nonequilibrium states as molecular conformations that are not at energy minima, which is different from the terminology in statistical mechanics. By adding random noise to each atomic coordinate, unrealistic conformation $\hat{{X}}$ can be obtained which has higher energy than the sampled conformation ${X}$. Driven by this, we assume $p(\hat{{X}}; {V})$ is approximated by a Gaussian distribution $q(\hat{{X}}; {X}, {V}) = \mathcal{N}({{X}}, \sigma  I_{3N})$ centered at ${X}$. Following the assumption, the force field is proportional to the perturbation noise for a given variance $\sigma$ as shown in Eq.~\ref{eq:denoise}.
\begin{equation}
    \nabla_{\hat{X}} \log p(\hat{X}; {V}) \approx \nabla_{\hat{X}} \log q(\hat{{X}}; {X}, {V}) \propto \frac{\hat{{X}} - {X}}{\sigma^2} \propto {\mathcal{E}}. 
    \label{eq:denoise}
\end{equation}
Therefore, training a GNN to match the perturbation noise is equivalent to learning a pseudo-force field when assuming the probability of atomic positions around a sampled conformation follows a Gaussian distribution. In our case, the temperature T in Boltzmann distribution is considered as a fictitious term when approximated with Gaussian, meaning $T$ is not explicitly included in the denoising pretraining. Such a simplification helps make use of data sampled from normal mode or torsional sampling that does not include temperature. Also, it is worth noting that temperature term $T$ in the Boltzmann distribution of a molecular system can be related to the selection of the optimal standard deviation $\sigma$ of the Gaussian distribution of random noise in pretraining. Investigation of data and pretraining strategies that connect $\sigma$ with $T$ is out of the scope of this work but could be an interesting direction to explore. 

It should be noted that though Gaussian distribution can be a good approximation of the distribution of states around a local minimum, it may fail for states with high energies. Such an assumption is related to the pretraining dataset and the selection of the standard deviation $\sigma$ of the Gaussian noise added to atomic positions. The pretraining molecular conformations should be physically reasonable and not overly distorted. Consequently, adding noise to these conformations leads to higher energy states, ensuring that the denoise pretraining process enables GNNs to learn a score function that leads to lower energy conformations. Details regarding the pretraining dataset can be found in section 3.1. Moreover, the selection of the standard deviation ($\sigma$) for the Gaussian noise affects the pretraining. When excessive noise is added, it distorts the conformation to such an extent that the denoise component struggles to accurately recover the original conformation with lower energy. On the other hand, if the perturbation is too small, the change of molecular energies can be trivial and it is hard for GNNs to learn meaning representations in pretraining. A detailed investigation of how $\sigma$ affects the performance of neural potential can be found in section 3.6. 


\subsection{Prediction of Noise}

To predict the noise given the perturbed molecular conformation, the GNN models are expected to output vectors instead of a single scalar. For SchNet and SE(3)-Transformer that lack the explicit modeling of vectorial features, models first output the predicted energy $\hat{E}$ (a scalar). The negative gradient with respect to the perturbed atomic coordinates $\hat{X}$ is calculated to evaluate the artificial noise, such that $\hat{\mathcal{E}} = - \nabla_{\hat{X}} \hat{E}$. On the other hand, EGNN and TorchMD-Net directly track the vectorial features which can be leveraged for noise prediction. In EGNN, the noise is evaluated as $\hat{\mathcal{E}} = X^{(T)} - \hat{X}$, where $X^{(T)}$ is the updated coordinates after $T$ message-passing layers. In TorchMD-Net, we adapt the gated equivariant block \cite{schutt2021equivariant} that maps the updated vectorial features $v^{(T)} \in \mathbb{R}^{N \times 3 \times F}$ and node features $h^{(T)} \in \mathbb{R}^{N \times F}$ after $T$ layers to the noise $\hat{\mathcal{E}} \in \mathbb{R}^{N \times 3}$ such that $\hat{\mathcal{E}} = \texttt{GatedEquivariant}(v^{(T)}, h^{(T)})$, which still keeps the equivariance.


\section{Results and Discussion}

\subsection{Datasets}

To evaluate the performance of denoise pretraining strategies on molecular potential predictions, five datasets containing various nonequilibrium molecular conformations with DFT-calculated energies are investigated as listed in Table~\ref{tb:dataset}. ANI-1 \cite{smith2017ani} selects a subset of GDB-17 \cite{ruddigkeit2012enumeration}, samples more than 24 million conformations via normal mode sampling, and calculates DFT total energy. ANI-1x \cite{smith2020ani} extends ANI-1 by obtaining over 5 million new conformations through an active learning algorithm \cite{smith2018less}. 

\begin{table}[t!]
  \centering
  \footnotesize
  \begin{tabular}{lrrrrll}
    \toprule
    Dataset & \# Mol. & \# Conf. & \# Ele. & \# Atoms & Molecule types & Usage \\
    \midrule
    ANI-1 \cite{smith2017ani}       & 57,462 & 24,687,809 & 4  & 2$\sim$26 & Small molecules & PT \& FT \\
    \midrule[0.1pt]
    ANI-1x \cite{smith2020ani}      & 63,865 & 5,496,771  & 4  & 2$\sim$63 & Small molecules & PT \& FT \\
    \midrule[0.1pt]
    ISO17 \cite{schutt2017schnet}   & 129    & 645,000    & 3  & 19 & Isomers of     C\textsubscript{7}O\textsubscript{2}H\textsubscript{10} & FT \\
    \midrule[0.1pt]
    SPICE \cite{eastman2023spice}   & 19,238 & 1,132,808  & 15  & 3$\sim$50 & Small molecules, dimers,  & FT \\
    & & & & & dipeptides, solvated amino acids & \\
    \midrule[0.1pt]
    MD22 \cite{chmiela2023accurate} & 7 & 223,422 & 4  & 42$\sim$370 & Proteins, lipids, carbohydrates, & FT \\
    & & & & & nucleic acids, supramolecules & \\
    \bottomrule
  \end{tabular}
\caption{Summary of datasets, including the number of molecules, number of conformations, number of elements, number of atoms per molecule, molecule types, and whether each dataset is used for pretraining (PT) and fine-tuning (FT).}
\label{tb:dataset}
\end{table}

Besides ANI-1 and ANI-1x, other datasets concerning molecular energetic predictions are studied. ISO17 \cite{schutt2017schnet} selects \textit{ab initio} molecular dynamics (AIMD) trajectories of molecular isomers with a fixed composition of atoms (i.e., C\textsubscript{7}O\textsubscript{2}H\textsubscript{10}), where each molecule contains 5,000 conformations with DFT-calculated energy and in total 129 molecules are included. Unlike the previous datasets which only investigate small organic molecules, MD22 \cite{chmiela2023accurate} and SPICE \cite{eastman2023spice} cover a wider variety of molecule types. In particular, MD22 includes the AIMD trajectories of proteins, carbohydrates, nucleic acids, and supramolecules (i.e., buckyball catcher and nanotube). Most molecular systems in MD22 contain more atoms than ANI-1 and ANI-1x and more details of MD22 can be found in Supplementary Information S1. SPICE covers more than 1.1 million conformations and is constituted by different molecular systems including small molecules, dimers, dipeptides, and amino acids. While other datasets contain only H, C, N, and O, SPICE involves molecules with halogens and metals, adding up to 15 different elements. It also includes charged and polar molecules, which further broaden the chemical space it covers. 

In this study, we combine ANI-1 and ANI-1x as the pretraining dataset since they include various small organic molecules with different conformations. In pretraining, all conformations of each molecule are split into the train and validation sets by a ratio of 95\%/5\%. All datasets including ANI-1 and ANI-1x are benchmarked in fine-tuning for potential predictions. By this means, we investigate whether invariant or equivariant GNNs pretrained on small molecules generalize to other various molecular systems. During fine-tuning, we split the dataset based on the conformations of each molecule by a ratio of 80\%/10\%/10\% into the train, validation, and test sets for ANI-1, ANI-1x, MD22, and SPICE. Besides, we follow the splitting strategy reported in the original literature \cite{schutt2017schnet} for ISO17. 


\subsection{Experimental Settings}

During pretraining, each invariant or equivariant GNN is trained for 5 epochs with a maximal learning rate $2 \times 10^{-4}$ and zero weight decay. All models are pretrained on the combination of ANI-1 and ANI-1x and fine-tuned on each dataset separately. In pretraining, we employ the AdamW optimizer \cite{loshchilov2017decoupled} with the batch size 256, and a linear learning rate warmup with cosine decay \cite{loshchilov2017sgdr} is applied. During fine-tuning, the models are trained for 10 epochs on ANI-1 and ANI-1x while trained for 50 epochs on SPICE and ISO17. We apply different experimental settings for each molecule in MD22 since molecules greatly vary in the number of atoms. In comparison to pretrained models, we also train GNNs from scratch on each dataset following the same setting as their denoise pretrained counterparts. Detailed fine-tuning settings can be found in Supplementary Information S2. In fine-tuning, the parameters of pretrained message-passing layers are transferred. Atomic features $h^K$ from message-passing layers are first summed up and fed into a randomly initialized MLP to predict the energy of each molecular conformation. Both the pretrained layers and the MLP head are fine-tuned for potential predictions. Hyperparameters for each GNN are based on the recommendations from the original literature. More details of GNNs implemented in this work can be found in Supplementary Information S3. To evaluate the performance of the neural potentials, we report both the rooted mean square error (RMSE) and mean absolute error (MAE) with the DFT potentials. 


\subsection{Neural Potential Predictions}

To investigate the performance of denoise pretraining, we pretrain the invariant and equivariant GNN models on the combination of ANI-1 and ANI-1x and fine-tune the models on each dataset separately. Table~\ref{tb:ani1_ani1x} compares the molecular potential prediction results of pretraining and no pretraining on ANI-1 and ANI-1x. 

\begin{table}[t!]
  \centering
  \footnotesize
  \begin{tabular}{l|c|cc|cc}
    \toprule
    & & \multicolumn{2}{c|}{ANI-1 } & \multicolumn{2}{c}{ANI-1x} \\
    \midrule
    Model & Pretrain & RMSE  & MAE & RMSE & MAE \\
    & & (kcal/mol) & (kcal/mol) & (kcal/mol) & (kcal/mol) \\
    \midrule
    SchNet &            & 2.03 & 1.33 & 9.29 & 5.95 \\
    SchNet & \checkmark & 1.50 & 1.00 & 5.53 & 3.80 \\
    \midrule
    SE(3)-Transformer &            & 5.82 & 4.14 & 17.67 & 11.89 \\
    SE(3)-Transformer & \checkmark & 5.12 & 3.64 & 13.76 & 9.50 \\
    \midrule
    EGNN &            & 2.35 & 1.63 & 7.35 & 5.30 \\
    EGNN & \checkmark & 1.66 & 1.14 & 4.94 & 3.49 \\
    \midrule
    TorchMD-Net &            & 0.60 & 0.39 & 2.27 & 1.50 \\
    TorchMD-Net & \checkmark & 0.49 & 0.32 & 1.54 & 1.01 \\
    \midrule
    Avg. improve & & \textcolor{olive}{-21.5\%} & \textcolor{olive}{-21.2\%} & \textcolor{olive}{-31.9\%} & \textcolor{olive}{-30.9\%} \\
    \bottomrule
  \end{tabular}
\caption{Performance of different invariant and equivariant GNNs on ANI-1 and ANI-1x with and without denoise pretraining.}
\label{tb:ani1_ani1x}
\end{table}
Our experiments demonstrate that the proposed denoise pretraining approach significantly improves the accuracy of GNNs for molecular potential predictions. Compared to their non-pretrained counterparts, the pretrained models achieve an average decrease of 21.5\% and 21.2\% in RMSE and MAE on ANI-1, and an average decrease of 31.9\% and 30.9\% in RMSE and MAE on ANI-1x. Furthermore, results also validate that the denoise pretraining method is model-agnostic, as it significantly improves the performance of four different GNNs, including SchNet, SE(3)-Transformer, EGNN, and TorchMD-Net, on both neural potential datasets. For instance, the invariant SchNet, as well as equivariant EGNN and TorchMD-Net gain more than 32\% improvement in both RMSE and MAE on ANI-1x. Especially, denoise pretraining boosts the performance of less computationally expensive GNN models like SchNet and EGNN. EGNN benefits the most from pretraining when benchmarked on both ANI-1 and ANI-1x. This elucidates that denoise pretraining can improve the performance of simple and cheap GNN models, which may mitigate the efforts of designing sophisticated equivariant architectures. This can be essential in applying neural potentials to real-world simulations since the efficiency of potential calculations is substantially important. Details of computational efficiency for each GNN model can be found in Supplementary Information S4. Also, the investigation of fine-tuning for more epochs can be found in Supplementary Information S5. These results demonstrate that denoising pretraining of invariant and equivariant GNNs on the molecular potential datasets can directly boost prediction accuracy. 


\subsection{Transferability}

The results presented in the previous section demonstrate that denoise pretraining is highly effective for improving the accuracy of neural potentials when the downstream tasks and pretraining data cover similar chemical space (i.e., small molecules). However, real-world simulations often involve much larger and more complex molecular systems, making it challenging to obtain the energies of such systems using expensive \textit{ab initio} methods. Therefore, the ability to transfer GNN models pretrained on small molecules (with sufficient training data) to larger and more complex molecular systems (with limited training data) would be highly beneficial. To this end, we fine-tune the pretrained four GNNs, either invariant or equivariant, on three other molecular potential benchmarks (i.e., ISO17 \cite{schutt2017schnet}, SPICE \cite{eastman2023spice}, and MD22 \cite{chmiela2023accurate}), which cover different molecular systems from the pretraining datasets. 

\begin{figure}[t!]
    \centering
    \includegraphics[width=\textwidth, keepaspectratio=true]{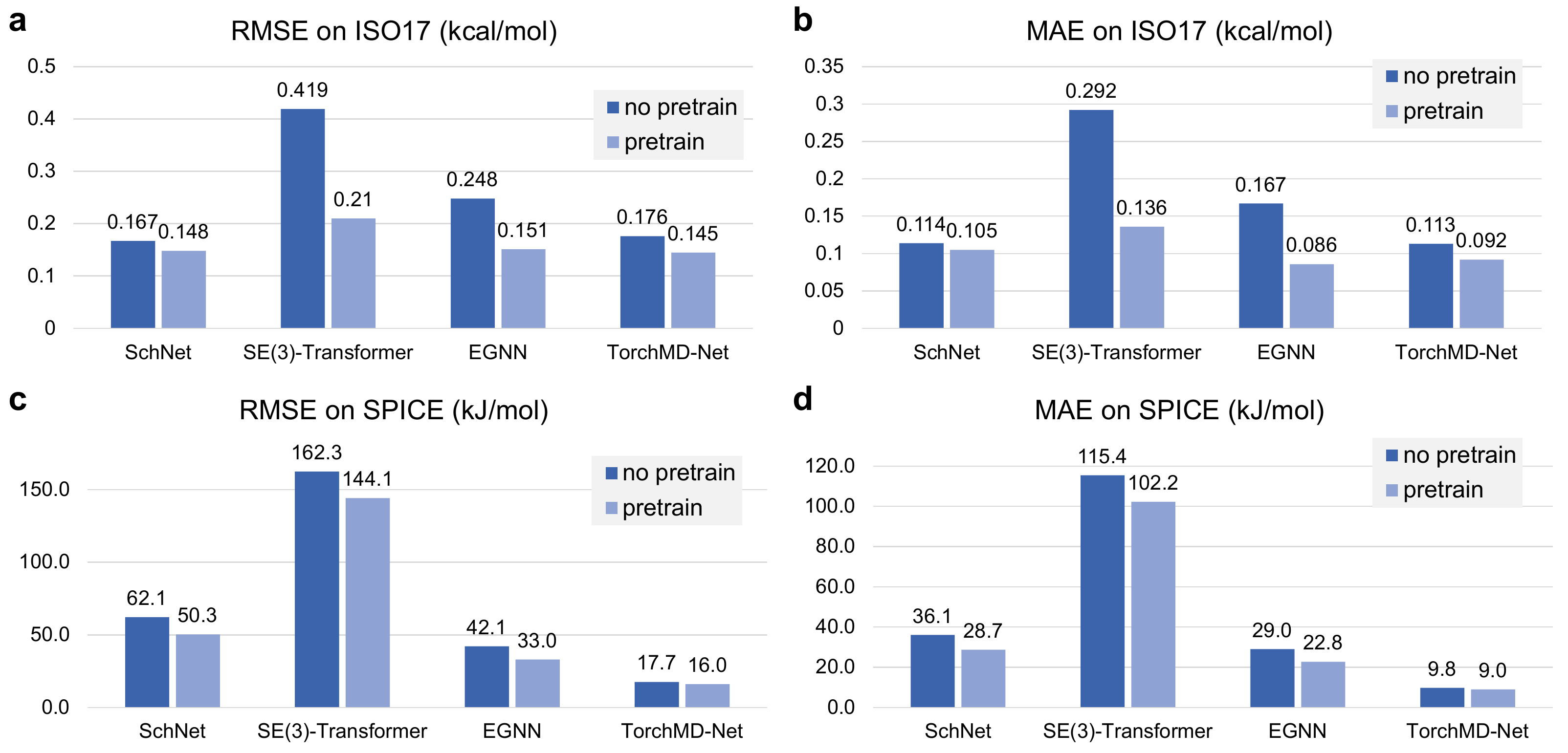}
    \caption{(a) Test RMSE of different GNN models on ISO17, (b) test MAE of different GNN models on ISO17, (c) test RMSE of different GNN models on SPICE, and (d) test MAE of different GNN models on SPICE.}
    \label{fig:iso17_spice}
\end{figure}

Figure~\ref{fig:iso17_spice}a and \ref{fig:iso17_spice}b compare the performance of GNN models with and without denoise pretraining on ISO17 containing isomers of C\textsubscript{7}O\textsubscript{2}H\textsubscript{10}. It is demonstrated that pretrained GNN models achieve better prediction accuracy when evaluated by both RMSE and MAE. Especially, pretraining significantly improves the performance of SE(3)-Transformer by approximately 50\%. Pretrained models are shown to be transferable to a specific molecular system containing fixed atoms. 

Figure~\ref{fig:iso17_spice}c and \ref{fig:iso17_spice}d show the performance of pretrained and non-pretrained GNNs on SPICE. SPICE includes halogen and metal elements, as well as charged and polar molecules that are absent from the pretraining dataset, making it a challenging benchmark. Nevertheless, pretrained GNNs still demonstrate superior performance than those trained from scratch. On average, pretrained GNNs show 15.3\% lower RMSE and MAE on SPICE. This elucidates that denoising pretraining benefits the transferability of GNN models even for different elements and electrostatic interactions. 

\begin{figure}[t!]
    \centering
    \includegraphics[width=\textwidth, keepaspectratio=true]{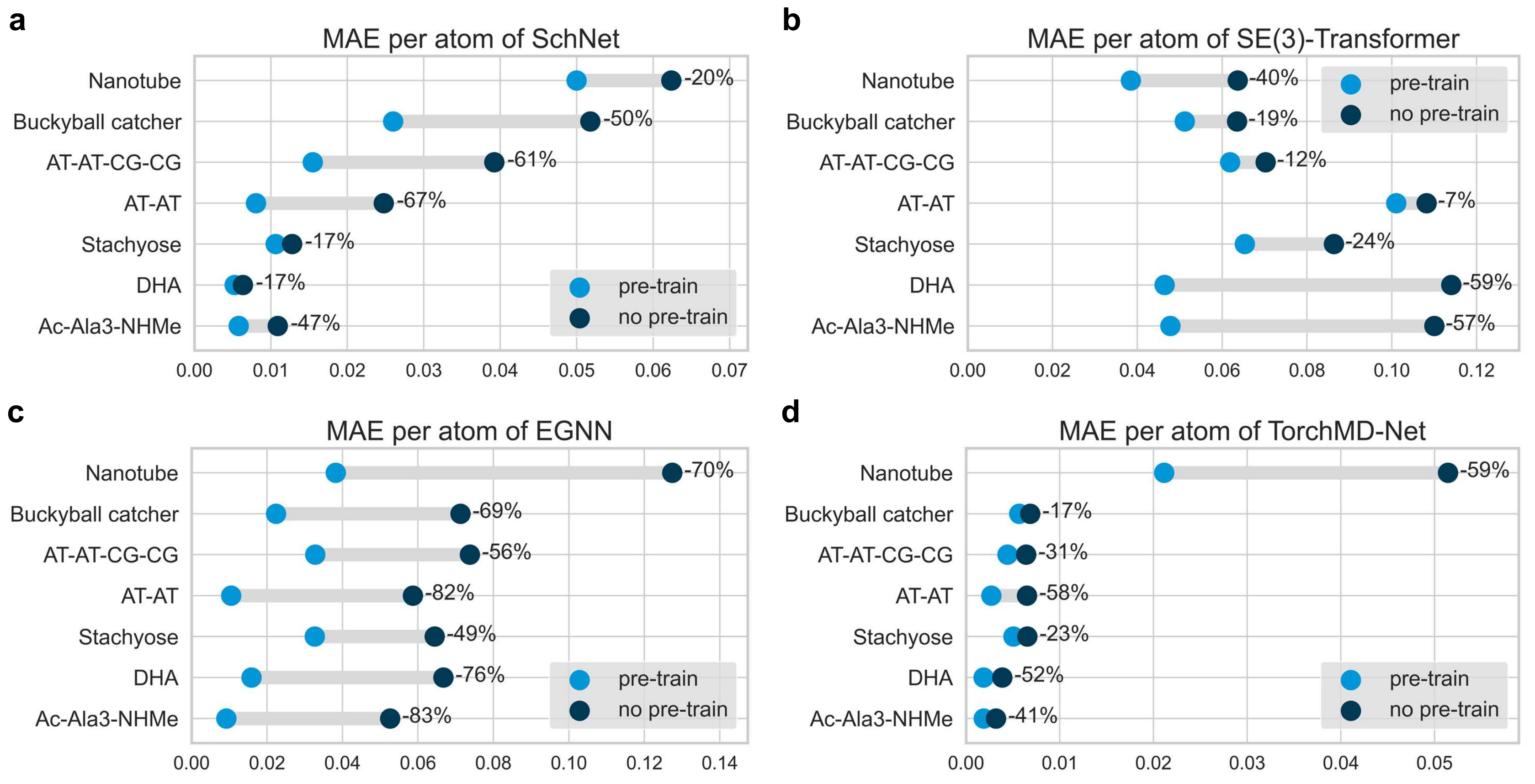}
    \caption{Test MAE per atom (kcal/mol) of (a) SchNet (b) SE(3)-Transformer (c) EGNN (d) TorchMD-Net on MD22 with and without denoise pretraining.}
    \label{fig:md22}
\end{figure}

Lastly, we fine-tune the pretrained models on MD22 which consists of larger molecular systems with numbers of atoms up to 370. This provides an appropriate benchmark to validate the transferability of GNN models pretrained on small molecules. Test MAE of the four GNN models on each molecule data is shown in Figure~\ref{fig:md22}. On all the different molecules in MD22, pretraining boosts the performance of neural potential predictions. It should be noted that MD22 contains a limited number of data compared with other benchmarks, especially for large molecules like the buckyball catcher (6,102 data) and double-walled nanotube (5,032 data). Pretrained SchNet, SE(3)-Transformer, EGNN, and TorchMD-Net fine-tuned on buckyball catcher and double-walled nanotube show average improvements of 39.3\% and 47.3\%, respectively. Also, among all the four GNNs, EGNN reaches the most significant improvement. Compared with SchNet and TorchMD-Net, EGNN does not include sophisticated architectural designs like interaction and update layers. The pretraining technique greatly improves the potential of expressiveness of simple equivariant GNN models and achieves competitive performance. These results further validate the effectiveness of denoise pretraining such that GNNs pretrained on small molecules can be transferred to large and complex systems for molecular potential predictions. Further investigation about the uncertainty of neural potential predictions in fine-tuning can be found in Supplementary Information S6. 

Overall, our experiments suggest that the proposed pretraining method can improve the accuracy and transferability of GNN models across diverse molecular systems, making it a promising method for predicting the energies of complex molecular systems with limited training data.


\subsection{Data Efficiency}

\begin{figure}[t!]
    \centering
    \includegraphics[width=\textwidth, keepaspectratio=true]{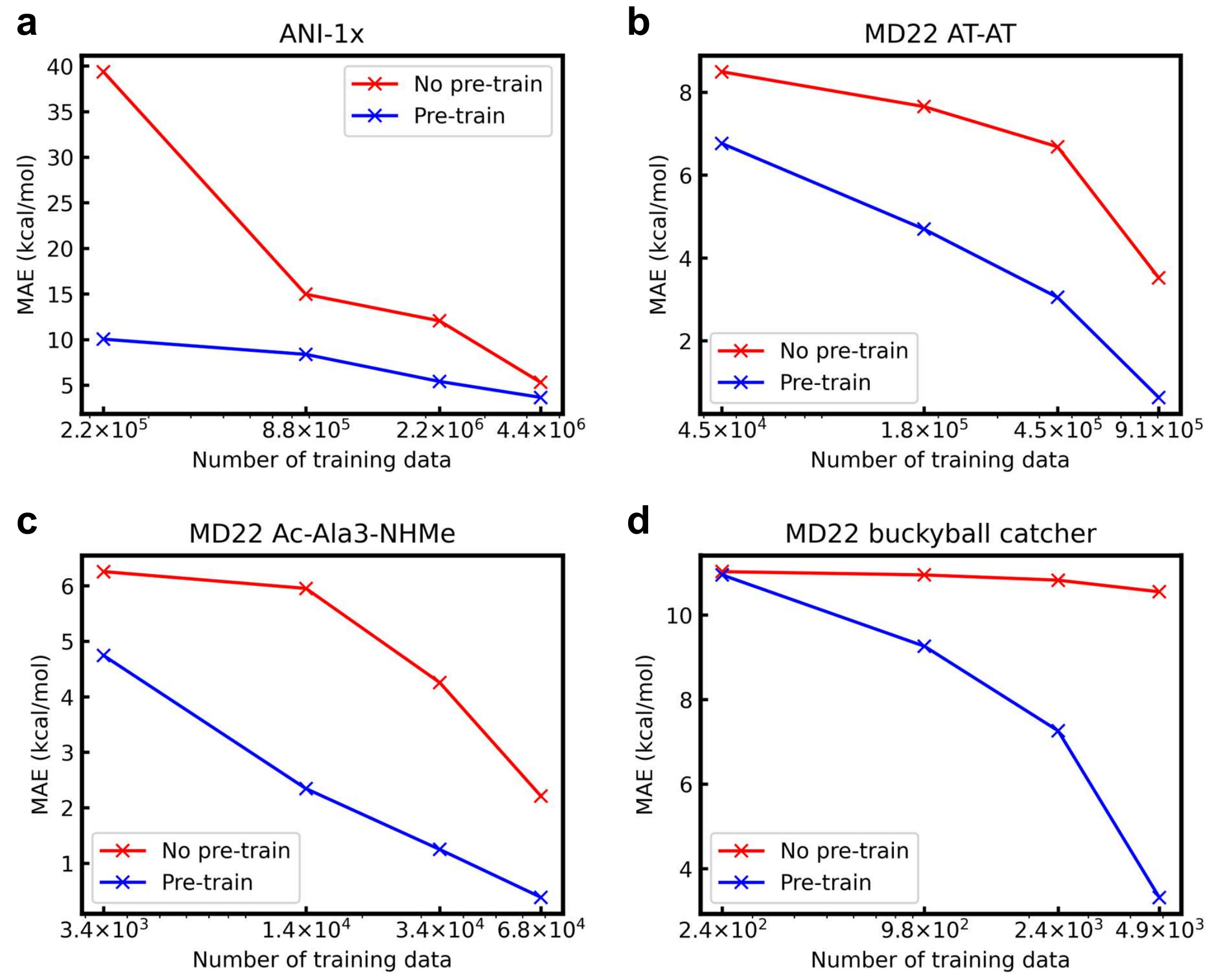}
    \caption{Test MAE of EGNN on (a) ANI-1x (b) MD22 AT-AT (c) MD22 Ac-Ala3-NHMe, and (d) MD22 buckyball catcher with different training data sizes.}
    \label{fig:data_size}
\end{figure}

To further evaluate the benefits of denoise pretraining for molecular potential predictions, we train GNNs with different dataset sizes. As shown in Figure~\ref{fig:data_size}, we compare the performance of pretrained and non-pretrained EGNN on ANI-1x as well as AT-AT, Ac-Ala3-NHMe, and buckyball catcher in MD22 with different numbers of data. Specifically, after splitting each dataset into the training, validation, and test sets, subsets that compose $\{5\%, 20\%, 50\%, 100\%\}$ data of the training set are sampled. It is exhibited that pretrained GNNs perform better than non-pretrained GNNs with the same number of training data. For instance, with only 5\% of the total training set, pretrained EGNN achieves an MAE of 10.04, which is much lower than the non-pretrained counterpart with an MAE of 39.37. Such performance is even better than non-pretrained EGNN trained on 10 times more training data. Besides, as shown in Figure~\ref{fig:data_size}d, both EGNNs with and without pretraining perform poorly on 5\% of buckyball catcher training data, which is less than 250. However, the accuracy of EGNN without pretraining barely improves even when trained on more buckyball catcher data, since the total training data is still limited to less than 5,000. The collection of molecular potentials via QM methods can be very expensive for such large and complex systems. On the other hand, when increasing numbers of data are fed, pretrained EGNN demonstrates improving performance. When using all training data, pretrained EGNN achieves a more than 68\% lower MAE compared with no pretraining. It is concluded that pretrained GNN models are more data efficient than models trained from scratch in achieving rival potential prediction performance. This is especially valuable when applying neural potentials for large and complex molecular systems with limited training data. 

\subsection{Selection of Noise}

\begin{figure}[t!]
    \centering
    \includegraphics[width=0.6\textwidth, keepaspectratio=true]{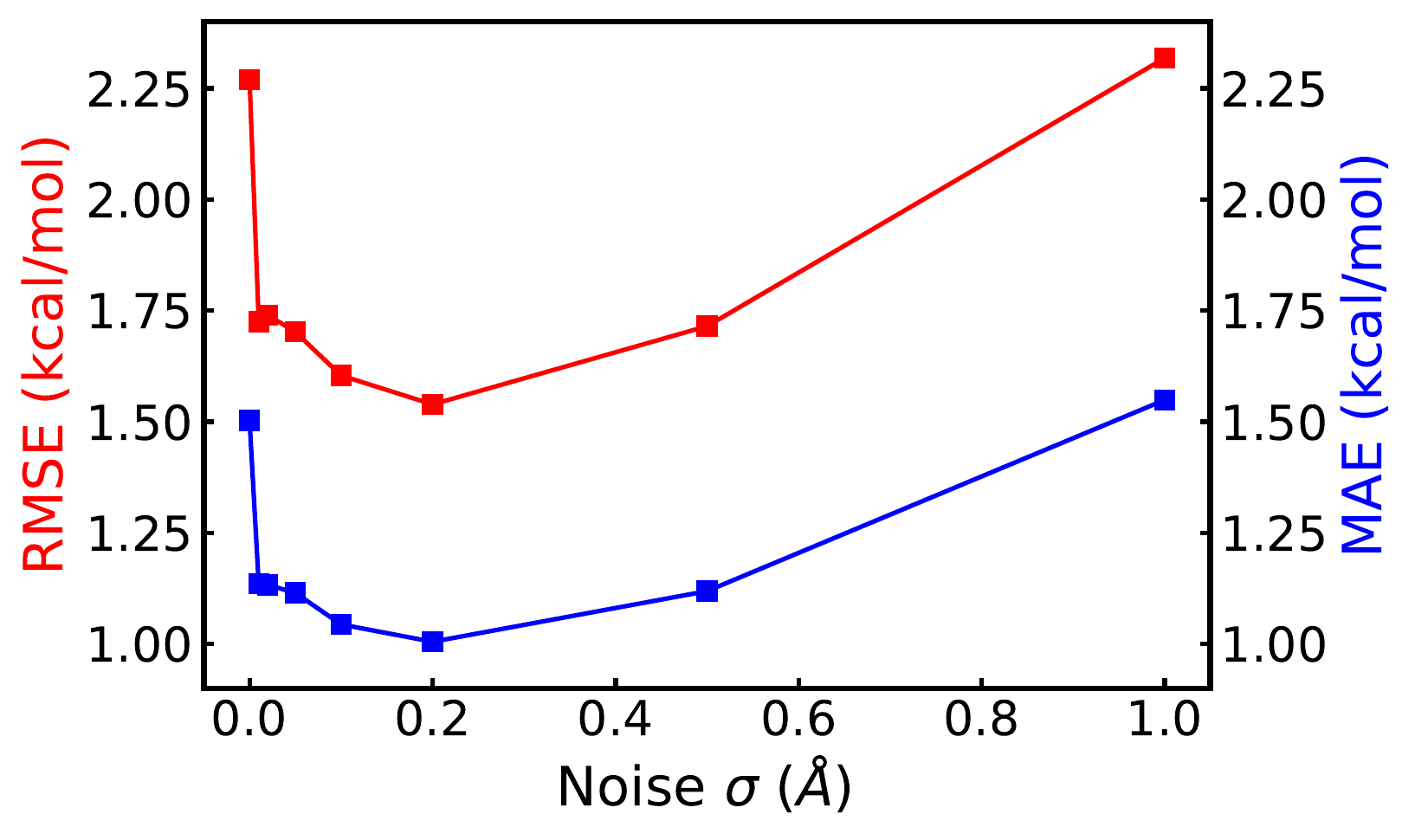}
    \caption{Influence of the standard deviation $\sigma$ of pretraining noise on the performance of neural potentials. Both test RMSE and MAE of TorchMD-Net on ANI-1x are shown.}
    \label{fig:noise}
\end{figure}

The standard deviation $\sigma$ in the Gaussian distribution of the noise to perturb molecular conformations affects the performance of pretrained equivariant GNNs. Small perturbation noise can be too trivial for the model to predict, while large noise may break the molecular conformations and fail to learn useful information for molecular simulations. To select the suitable $\sigma$, we enumerate different $\sigma$ values $\{ 0.01, 0.02, 0.05, 0.1, 0.2, 0.5, 1.0 \} \angstrom$. Fig.~\ref{fig:noise} illustrates the performance of pretrained TorchMD-Net on ANI-1x with different noise scales. TorchMD-Net without pretraining is included as $\sigma=0.0 \angstrom$ for comparison. As shown, pretrained models achieve the best performance with noise $\sigma=0.2 \angstrom$ on both RMSE and MAE of the predicted energies. Also, pretraining with other noise scales from 0.01 to 0.5 effectively boost neural potential predictions. However, when large noise is applied (e.g., $\sigma=1.0 \angstrom$), pretraining harms the potential prediction since the perturbed conformations are far away from the local minimum of the sampled conformations. This could be due to that large noise sabotages the original conformations which breaks the assumption in Section \ref{sec:denoise} and GNNs fail to learn meaning representations concerning potentials. Based on the experimental results, we select $\sigma=0.2$ in other experiments. 

\section{Conclusions}

To summarize, our proposed denoise pretraining method for invariant and equivariant graph neural networks (GNNs) on nonequilibrium molecular conformations enables more accurate and transferable neural potential predictions. Our rigorous experiments across multiple benchmarks demonstrate that pretraining significantly improves the accuracy of neural potentials. Furthermore, GNNs pretrained on small molecules through denoising exhibit superior transferability and data efficiency for diverse molecular systems, including different elements, polar molecules, biomolecules, and larger systems. This transferability is particularly valuable for building neural potential models on larger and more complex systems where sufficient data is often challenging to obtain. Notably, the model-agnostic nature of our pretraining method is confirmed by the performance improvements observed across different invariant and equivariant GNN models, including SchNet, SE(3)-Transformer, EGNN, and TorchMD-Net. Our proposed denoise pretraining method thus paves the way for improving neural potential predictions and holds great potential for broader applications in molecular simulations.

\begin{acknowledgement}

The author thanks the Department of Mechanical Engineering Department at Carnegie Mellon University for the start-up fund to support the work.

\end{acknowledgement}

\section*{Data and Code Availability}

The code as well as the data used in this work can be found on the GitHub repository: \url{https://github.com/yuyangw/Denoise-Pretrain-ML-Potential}.

\begin{suppinfo}

Summary of MD22, detailed implementations of GNNs, details of fine-tuning settings, computational efficiency for each GNN, investigation of fine-tuning epochs, and uncertainty in fine-tuning are included in the Supplementary Information. \url{https://pubs.acs.org/doi/suppl/10.1021/acs.jctc.3c00289/suppl_file/ct3c00289_si_001.pdf}

\end{suppinfo}


\begin{mcitethebibliography}{80}
\providecommand*\natexlab[1]{#1}
\providecommand*\mciteSetBstSublistMode[1]{}
\providecommand*\mciteSetBstMaxWidthForm[2]{}
\providecommand*\mciteBstWouldAddEndPuncttrue
  {\def\EndOfBibitem{\unskip.}}
\providecommand*\mciteBstWouldAddEndPunctfalse
  {\let\EndOfBibitem\relax}
\providecommand*\mciteSetBstMidEndSepPunct[3]{}
\providecommand*\mciteSetBstSublistLabelBeginEnd[3]{}
\providecommand*\EndOfBibitem{}
\mciteSetBstSublistMode{f}
\mciteSetBstMaxWidthForm{subitem}{(\alph{mcitesubitemcount})}
\mciteSetBstSublistLabelBeginEnd
  {\mcitemaxwidthsubitemform\space}
  {\relax}
  {\relax}

\bibitem[Monkhorst(1977)]{monkhorst1977calculation}
Monkhorst,~H.~J. Calculation of properties with the coupled-cluster method.
  \emph{International Journal of Quantum Chemistry} \textbf{1977}, \emph{12},
  421--432\relax
\mciteBstWouldAddEndPuncttrue
\mciteSetBstMidEndSepPunct{\mcitedefaultmidpunct}
{\mcitedefaultendpunct}{\mcitedefaultseppunct}\relax
\EndOfBibitem
\bibitem[Hirata \latin{et~al.}(2004)Hirata, Podeszwa, Tobita, and
  Bartlett]{hirata2004coupled}
Hirata,~S.; Podeszwa,~R.; Tobita,~M.; Bartlett,~R.~J. Coupled-cluster singles
  and doubles for extended systems. \emph{The Journal of chemical physics}
  \textbf{2004}, \emph{120}, 2581--2592\relax
\mciteBstWouldAddEndPuncttrue
\mciteSetBstMidEndSepPunct{\mcitedefaultmidpunct}
{\mcitedefaultendpunct}{\mcitedefaultseppunct}\relax
\EndOfBibitem
\bibitem[Cremer(2011)]{cremer2011moller}
Cremer,~D. M{\o}ller--Plesset perturbation theory: from small molecule methods
  to methods for thousands of atoms. \emph{Wiley Interdisciplinary Reviews:
  Computational Molecular Science} \textbf{2011}, \emph{1}, 509--530\relax
\mciteBstWouldAddEndPuncttrue
\mciteSetBstMidEndSepPunct{\mcitedefaultmidpunct}
{\mcitedefaultendpunct}{\mcitedefaultseppunct}\relax
\EndOfBibitem
\bibitem[Levy(1979)]{levy1979universal}
Levy,~M. Universal variational functionals of electron densities, first-order
  density matrices, and natural spin-orbitals and solution of the
  v-representability problem. \emph{Proceedings of the National Academy of
  Sciences} \textbf{1979}, \emph{76}, 6062--6065\relax
\mciteBstWouldAddEndPuncttrue
\mciteSetBstMidEndSepPunct{\mcitedefaultmidpunct}
{\mcitedefaultendpunct}{\mcitedefaultseppunct}\relax
\EndOfBibitem
\bibitem[Thanthiriwatte \latin{et~al.}(2011)Thanthiriwatte, Hohenstein, Burns,
  and Sherrill]{thanthiriwatte2011assessment}
Thanthiriwatte,~K.~S.; Hohenstein,~E.~G.; Burns,~L.~A.; Sherrill,~C.~D.
  Assessment of the performance of DFT and DFT-D methods for describing
  distance dependence of hydrogen-bonded interactions. \emph{Journal of
  Chemical Theory and Computation} \textbf{2011}, \emph{7}, 88--96\relax
\mciteBstWouldAddEndPuncttrue
\mciteSetBstMidEndSepPunct{\mcitedefaultmidpunct}
{\mcitedefaultendpunct}{\mcitedefaultseppunct}\relax
\EndOfBibitem
\bibitem[Mark and Nilsson(2001)Mark, and Nilsson]{mark2001structure}
Mark,~P.; Nilsson,~L. Structure and dynamics of the TIP3P, SPC, and SPC/E water
  models at 298 K. \emph{The Journal of Physical Chemistry A} \textbf{2001},
  \emph{105}, 9954--9960\relax
\mciteBstWouldAddEndPuncttrue
\mciteSetBstMidEndSepPunct{\mcitedefaultmidpunct}
{\mcitedefaultendpunct}{\mcitedefaultseppunct}\relax
\EndOfBibitem
\bibitem[Halgren(1996)]{halgren1996merck}
Halgren,~T.~A. Merck molecular force field. I. Basis, form, scope,
  parameterization, and performance of MMFF94. \emph{Journal of computational
  chemistry} \textbf{1996}, \emph{17}, 490--519\relax
\mciteBstWouldAddEndPuncttrue
\mciteSetBstMidEndSepPunct{\mcitedefaultmidpunct}
{\mcitedefaultendpunct}{\mcitedefaultseppunct}\relax
\EndOfBibitem
\bibitem[Salomon-Ferrer \latin{et~al.}(2013)Salomon-Ferrer, Case, and
  Walker]{salomon2013overview}
Salomon-Ferrer,~R.; Case,~D.~A.; Walker,~R.~C. An overview of the Amber
  biomolecular simulation package. \emph{Wiley Interdisciplinary Reviews:
  Computational Molecular Science} \textbf{2013}, \emph{3}, 198--210\relax
\mciteBstWouldAddEndPuncttrue
\mciteSetBstMidEndSepPunct{\mcitedefaultmidpunct}
{\mcitedefaultendpunct}{\mcitedefaultseppunct}\relax
\EndOfBibitem
\bibitem[Huang and MacKerell~Jr(2013)Huang, and
  MacKerell~Jr]{huang2013charmm36}
Huang,~J.; MacKerell~Jr,~A.~D. CHARMM36 all-atom additive protein force field:
  Validation based on comparison to NMR data. \emph{Journal of computational
  chemistry} \textbf{2013}, \emph{34}, 2135--2145\relax
\mciteBstWouldAddEndPuncttrue
\mciteSetBstMidEndSepPunct{\mcitedefaultmidpunct}
{\mcitedefaultendpunct}{\mcitedefaultseppunct}\relax
\EndOfBibitem
\bibitem[Sun(1998)]{sun1998compass}
Sun,~H. COMPASS: an ab initio force-field optimized for condensed-phase
  applications overview with details on alkane and benzene compounds. \emph{The
  Journal of Physical Chemistry B} \textbf{1998}, \emph{102}, 7338--7364\relax
\mciteBstWouldAddEndPuncttrue
\mciteSetBstMidEndSepPunct{\mcitedefaultmidpunct}
{\mcitedefaultendpunct}{\mcitedefaultseppunct}\relax
\EndOfBibitem
\bibitem[Vitalini \latin{et~al.}(2015)Vitalini, Mey, No{\'e}, and
  Keller]{vitalini2015dynamic}
Vitalini,~F.; Mey,~A.~S.; No{\'e},~F.; Keller,~B.~G. Dynamic properties of
  force fields. \emph{The Journal of Chemical Physics} \textbf{2015},
  \emph{142}, 02B611\_1\relax
\mciteBstWouldAddEndPuncttrue
\mciteSetBstMidEndSepPunct{\mcitedefaultmidpunct}
{\mcitedefaultendpunct}{\mcitedefaultseppunct}\relax
\EndOfBibitem
\bibitem[Harrison \latin{et~al.}(2018)Harrison, Schall, Maskey, Mikulski,
  Knippenberg, and Morrow]{harrison2018review}
Harrison,~J.~A.; Schall,~J.~D.; Maskey,~S.; Mikulski,~P.~T.;
  Knippenberg,~M.~T.; Morrow,~B.~H. Review of force fields and intermolecular
  potentials used in atomistic computational materials research. \emph{Applied
  Physics Reviews} \textbf{2018}, \emph{5}, 031104\relax
\mciteBstWouldAddEndPuncttrue
\mciteSetBstMidEndSepPunct{\mcitedefaultmidpunct}
{\mcitedefaultendpunct}{\mcitedefaultseppunct}\relax
\EndOfBibitem
\bibitem[Unke \latin{et~al.}(2020)Unke, Koner, Patra, K{\"a}ser, and
  Meuwly]{unke2020high}
Unke,~O.~T.; Koner,~D.; Patra,~S.; K{\"a}ser,~S.; Meuwly,~M. High-dimensional
  potential energy surfaces for molecular simulations: from empiricism to
  machine learning. \emph{Machine Learning: Science and Technology}
  \textbf{2020}, \emph{1}, 013001\relax
\mciteBstWouldAddEndPuncttrue
\mciteSetBstMidEndSepPunct{\mcitedefaultmidpunct}
{\mcitedefaultendpunct}{\mcitedefaultseppunct}\relax
\EndOfBibitem
\bibitem[Butler \latin{et~al.}(2018)Butler, Davies, Cartwright, Isayev, and
  Walsh]{butler2018machine}
Butler,~K.~T.; Davies,~D.~W.; Cartwright,~H.; Isayev,~O.; Walsh,~A. Machine
  learning for molecular and materials science. \emph{Nature} \textbf{2018},
  \emph{559}, 547--555\relax
\mciteBstWouldAddEndPuncttrue
\mciteSetBstMidEndSepPunct{\mcitedefaultmidpunct}
{\mcitedefaultendpunct}{\mcitedefaultseppunct}\relax
\EndOfBibitem
\bibitem[Unke \latin{et~al.}(2021)Unke, Chmiela, Sauceda, Gastegger, Poltavsky,
  Sch\"{u}tt, Tkatchenko, and M\"{u}ller]{unke2021machine}
Unke,~O.~T.; Chmiela,~S.; Sauceda,~H.~E.; Gastegger,~M.; Poltavsky,~I.;
  Sch\"{u}tt,~K.~T.; Tkatchenko,~A.; M\"{u}ller,~K.-R. Machine learning force
  fields. \emph{Chemical Reviews} \textbf{2021}, \emph{121}, 10142--10186\relax
\mciteBstWouldAddEndPuncttrue
\mciteSetBstMidEndSepPunct{\mcitedefaultmidpunct}
{\mcitedefaultendpunct}{\mcitedefaultseppunct}\relax
\EndOfBibitem
\bibitem[Li \latin{et~al.}(2022)Li, Meidani, Yadav, and
  Barati~Farimani]{li2022graph}
Li,~Z.; Meidani,~K.; Yadav,~P.; Barati~Farimani,~A. Graph neural networks
  accelerated molecular dynamics. \emph{The Journal of Chemical Physics}
  \textbf{2022}, \emph{156}, 144103\relax
\mciteBstWouldAddEndPuncttrue
\mciteSetBstMidEndSepPunct{\mcitedefaultmidpunct}
{\mcitedefaultendpunct}{\mcitedefaultseppunct}\relax
\EndOfBibitem
\bibitem[Fu \latin{et~al.}(2022)Fu, Wu, Wang, Xie, Keten, Gomez-Bombarelli, and
  Jaakkola]{fu2022forces}
Fu,~X.; Wu,~Z.; Wang,~W.; Xie,~T.; Keten,~S.; Gomez-Bombarelli,~R.;
  Jaakkola,~T. Forces are not enough: Benchmark and critical evaluation for
  machine learning force fields with molecular simulations. \emph{arXiv
  preprint arXiv:2210.07237} \textbf{2022}, \relax
\mciteBstWouldAddEndPunctfalse
\mciteSetBstMidEndSepPunct{\mcitedefaultmidpunct}
{}{\mcitedefaultseppunct}\relax
\EndOfBibitem
\bibitem[Atz \latin{et~al.}(2021)Atz, Grisoni, and Schneider]{atz2021geometric}
Atz,~K.; Grisoni,~F.; Schneider,~G. Geometric deep learning on molecular
  representations. \emph{Nature Machine Intelligence} \textbf{2021}, \emph{3},
  1023--1032\relax
\mciteBstWouldAddEndPuncttrue
\mciteSetBstMidEndSepPunct{\mcitedefaultmidpunct}
{\mcitedefaultendpunct}{\mcitedefaultseppunct}\relax
\EndOfBibitem
\bibitem[Chmiela \latin{et~al.}(2017)Chmiela, Tkatchenko, Sauceda, Poltavsky,
  Sch\"{u}tt, and M\"{u}ller]{chmiela2017machine}
Chmiela,~S.; Tkatchenko,~A.; Sauceda,~H.~E.; Poltavsky,~I.; Sch\"{u}tt,~K.~T.;
  M\"{u}ller,~K.-R. Machine learning of accurate energy-conserving molecular
  force fields. \emph{Science advances} \textbf{2017}, \emph{3}, e1603015\relax
\mciteBstWouldAddEndPuncttrue
\mciteSetBstMidEndSepPunct{\mcitedefaultmidpunct}
{\mcitedefaultendpunct}{\mcitedefaultseppunct}\relax
\EndOfBibitem
\bibitem[Chmiela \latin{et~al.}(2018)Chmiela, Sauceda, M{\"u}ller, and
  Tkatchenko]{chmiela2018towards}
Chmiela,~S.; Sauceda,~H.~E.; M{\"u}ller,~K.-R.; Tkatchenko,~A. Towards exact
  molecular dynamics simulations with machine-learned force fields.
  \emph{Nature communications} \textbf{2018}, \emph{9}, 1--10\relax
\mciteBstWouldAddEndPuncttrue
\mciteSetBstMidEndSepPunct{\mcitedefaultmidpunct}
{\mcitedefaultendpunct}{\mcitedefaultseppunct}\relax
\EndOfBibitem
\bibitem[Bart{\'o}k \latin{et~al.}(2010)Bart{\'o}k, Payne, Kondor, and
  Cs{\'a}nyi]{bartok2010gaussian}
Bart{\'o}k,~A.~P.; Payne,~M.~C.; Kondor,~R.; Cs{\'a}nyi,~G. Gaussian
  approximation potentials: The accuracy of quantum mechanics, without the
  electrons. \emph{Physical review letters} \textbf{2010}, \emph{104},
  136403\relax
\mciteBstWouldAddEndPuncttrue
\mciteSetBstMidEndSepPunct{\mcitedefaultmidpunct}
{\mcitedefaultendpunct}{\mcitedefaultseppunct}\relax
\EndOfBibitem
\bibitem[Bart{\'o}k \latin{et~al.}(2017)Bart{\'o}k, De, Poelking, Bernstein,
  Kermode, Cs{\'a}nyi, and Ceriotti]{bartok2017machine}
Bart{\'o}k,~A.~P.; De,~S.; Poelking,~C.; Bernstein,~N.; Kermode,~J.~R.;
  Cs{\'a}nyi,~G.; Ceriotti,~M. Machine learning unifies the modeling of
  materials and molecules. \emph{Science advances} \textbf{2017}, \emph{3},
  e1701816\relax
\mciteBstWouldAddEndPuncttrue
\mciteSetBstMidEndSepPunct{\mcitedefaultmidpunct}
{\mcitedefaultendpunct}{\mcitedefaultseppunct}\relax
\EndOfBibitem
\bibitem[Bart{\'o}k \latin{et~al.}(2013)Bart{\'o}k, Kondor, and
  Cs{\'a}nyi]{bartok2013representing}
Bart{\'o}k,~A.~P.; Kondor,~R.; Cs{\'a}nyi,~G. On representing chemical
  environments. \emph{Physical Review B} \textbf{2013}, \emph{87}, 184115\relax
\mciteBstWouldAddEndPuncttrue
\mciteSetBstMidEndSepPunct{\mcitedefaultmidpunct}
{\mcitedefaultendpunct}{\mcitedefaultseppunct}\relax
\EndOfBibitem
\bibitem[Grisafi \latin{et~al.}(2018)Grisafi, Wilkins, Cs{\'a}nyi, and
  Ceriotti]{grisafi2018symmetry}
Grisafi,~A.; Wilkins,~D.~M.; Cs{\'a}nyi,~G.; Ceriotti,~M. Symmetry-adapted
  machine learning for tensorial properties of atomistic systems.
  \emph{Physical review letters} \textbf{2018}, \emph{120}, 036002\relax
\mciteBstWouldAddEndPuncttrue
\mciteSetBstMidEndSepPunct{\mcitedefaultmidpunct}
{\mcitedefaultendpunct}{\mcitedefaultseppunct}\relax
\EndOfBibitem
\bibitem[Fedik \latin{et~al.}(2022)Fedik, Zubatyuk, Kulichenko, Lubbers, Smith,
  Nebgen, Messerly, Li, Boldyrev, Barros, \latin{et~al.}
  others]{fedik2022extending}
Fedik,~N.; Zubatyuk,~R.; Kulichenko,~M.; Lubbers,~N.; Smith,~J.~S.; Nebgen,~B.;
  Messerly,~R.; Li,~Y.~W.; Boldyrev,~A.~I.; Barros,~K., \latin{et~al.}
  Extending machine learning beyond interatomic potentials for predicting
  molecular properties. \emph{Nature Reviews Chemistry} \textbf{2022},
  \emph{6}, 653--672\relax
\mciteBstWouldAddEndPuncttrue
\mciteSetBstMidEndSepPunct{\mcitedefaultmidpunct}
{\mcitedefaultendpunct}{\mcitedefaultseppunct}\relax
\EndOfBibitem
\bibitem[Blank \latin{et~al.}(1995)Blank, Brown, Calhoun, and
  Doren]{blank1995neural}
Blank,~T.~B.; Brown,~S.~D.; Calhoun,~A.~W.; Doren,~D.~J. Neural network models
  of potential energy surfaces. \emph{The Journal of chemical physics}
  \textbf{1995}, \emph{103}, 4129--4137\relax
\mciteBstWouldAddEndPuncttrue
\mciteSetBstMidEndSepPunct{\mcitedefaultmidpunct}
{\mcitedefaultendpunct}{\mcitedefaultseppunct}\relax
\EndOfBibitem
\bibitem[Brown \latin{et~al.}(1996)Brown, Gibbs, and Clary]{brown1996combining}
Brown,~D.~F.; Gibbs,~M.~N.; Clary,~D.~C. Combining ab initio computations,
  neural networks, and diffusion Monte Carlo: An efficient method to treat
  weakly bound molecules. \emph{The Journal of chemical physics} \textbf{1996},
  \emph{105}, 7597--7604\relax
\mciteBstWouldAddEndPuncttrue
\mciteSetBstMidEndSepPunct{\mcitedefaultmidpunct}
{\mcitedefaultendpunct}{\mcitedefaultseppunct}\relax
\EndOfBibitem
\bibitem[Behler and Parrinello(2007)Behler, and
  Parrinello]{behler2007generalized}
Behler,~J.; Parrinello,~M. Generalized neural-network representation of
  high-dimensional potential-energy surfaces. \emph{Physical review letters}
  \textbf{2007}, \emph{98}, 146401\relax
\mciteBstWouldAddEndPuncttrue
\mciteSetBstMidEndSepPunct{\mcitedefaultmidpunct}
{\mcitedefaultendpunct}{\mcitedefaultseppunct}\relax
\EndOfBibitem
\bibitem[Behler(2011)]{behler2011atom}
Behler,~J. Atom-centered symmetry functions for constructing high-dimensional
  neural network potentials. \emph{The Journal of chemical physics}
  \textbf{2011}, \emph{134}, 074106\relax
\mciteBstWouldAddEndPuncttrue
\mciteSetBstMidEndSepPunct{\mcitedefaultmidpunct}
{\mcitedefaultendpunct}{\mcitedefaultseppunct}\relax
\EndOfBibitem
\bibitem[Smith \latin{et~al.}(2017)Smith, Isayev, and Roitberg]{smith2017ani}
Smith,~J.~S.; Isayev,~O.; Roitberg,~A.~E. ANI-1: an extensible neural network
  potential with DFT accuracy at force field computational cost. \emph{Chemical
  science} \textbf{2017}, \emph{8}, 3192--3203\relax
\mciteBstWouldAddEndPuncttrue
\mciteSetBstMidEndSepPunct{\mcitedefaultmidpunct}
{\mcitedefaultendpunct}{\mcitedefaultseppunct}\relax
\EndOfBibitem
\bibitem[Smith \latin{et~al.}(2018)Smith, Nebgen, Lubbers, Isayev, and
  Roitberg]{smith2018less}
Smith,~J.~S.; Nebgen,~B.; Lubbers,~N.; Isayev,~O.; Roitberg,~A.~E. Less is
  more: Sampling chemical space with active learning. \emph{The Journal of
  chemical physics} \textbf{2018}, \emph{148}, 241733\relax
\mciteBstWouldAddEndPuncttrue
\mciteSetBstMidEndSepPunct{\mcitedefaultmidpunct}
{\mcitedefaultendpunct}{\mcitedefaultseppunct}\relax
\EndOfBibitem
\bibitem[Devereux \latin{et~al.}(2020)Devereux, Smith, Huddleston, Barros,
  Zubatyuk, Isayev, and Roitberg]{devereux2020extending}
Devereux,~C.; Smith,~J.~S.; Huddleston,~K.~K.; Barros,~K.; Zubatyuk,~R.;
  Isayev,~O.; Roitberg,~A.~E. Extending the applicability of the ANI deep
  learning molecular potential to sulfur and halogens. \emph{Journal of
  Chemical Theory and Computation} \textbf{2020}, \emph{16}, 4192--4202\relax
\mciteBstWouldAddEndPuncttrue
\mciteSetBstMidEndSepPunct{\mcitedefaultmidpunct}
{\mcitedefaultendpunct}{\mcitedefaultseppunct}\relax
\EndOfBibitem
\bibitem[Yao \latin{et~al.}(2018)Yao, Herr, Toth, Mckintyre, and
  Parkhill]{yao2018tensormol}
Yao,~K.; Herr,~J.~E.; Toth,~D.~W.; Mckintyre,~R.; Parkhill,~J. The
  TensorMol-0.1 model chemistry: a neural network augmented with long-range
  physics. \emph{Chemical science} \textbf{2018}, \emph{9}, 2261--2269\relax
\mciteBstWouldAddEndPuncttrue
\mciteSetBstMidEndSepPunct{\mcitedefaultmidpunct}
{\mcitedefaultendpunct}{\mcitedefaultseppunct}\relax
\EndOfBibitem
\bibitem[Gilmer \latin{et~al.}(2017)Gilmer, Schoenholz, Riley, Vinyals, and
  Dahl]{gilmer2017neural}
Gilmer,~J.; Schoenholz,~S.~S.; Riley,~P.~F.; Vinyals,~O.; Dahl,~G.~E. Neural
  message passing for quantum chemistry. International conference on machine
  learning. 2017; pp 1263--1272\relax
\mciteBstWouldAddEndPuncttrue
\mciteSetBstMidEndSepPunct{\mcitedefaultmidpunct}
{\mcitedefaultendpunct}{\mcitedefaultseppunct}\relax
\EndOfBibitem
\bibitem[Sch\"{u}tt \latin{et~al.}(2018)Sch\"{u}tt, Sauceda, Kindermans,
  Tkatchenko, and M\"{u}ller]{schutt2018schnet}
Sch\"{u}tt,~K.~T.; Sauceda,~H.~E.; Kindermans,~P.-J.; Tkatchenko,~A.;
  M\"{u}ller,~K.-R. Schnet--a deep learning architecture for molecules and
  materials. \emph{The Journal of Chemical Physics} \textbf{2018}, \emph{148},
  241722\relax
\mciteBstWouldAddEndPuncttrue
\mciteSetBstMidEndSepPunct{\mcitedefaultmidpunct}
{\mcitedefaultendpunct}{\mcitedefaultseppunct}\relax
\EndOfBibitem
\bibitem[Gasteiger \latin{et~al.}(2020)Gasteiger, Groß, and
  Günnemann]{Gasteiger2020Directional}
Gasteiger,~J.; Groß,~J.; Günnemann,~S. Directional Message Passing for
  Molecular Graphs. International Conference on Learning Representations.
  2020\relax
\mciteBstWouldAddEndPuncttrue
\mciteSetBstMidEndSepPunct{\mcitedefaultmidpunct}
{\mcitedefaultendpunct}{\mcitedefaultseppunct}\relax
\EndOfBibitem
\bibitem[Unke and Meuwly(2019)Unke, and Meuwly]{unke2019physnet}
Unke,~O.~T.; Meuwly,~M. PhysNet: A neural network for predicting energies,
  forces, dipole moments, and partial charges. \emph{Journal of chemical theory
  and computation} \textbf{2019}, \emph{15}, 3678--3693\relax
\mciteBstWouldAddEndPuncttrue
\mciteSetBstMidEndSepPunct{\mcitedefaultmidpunct}
{\mcitedefaultendpunct}{\mcitedefaultseppunct}\relax
\EndOfBibitem
\bibitem[Lubbers \latin{et~al.}(2018)Lubbers, Smith, and
  Barros]{lubbers2018hierarchical}
Lubbers,~N.; Smith,~J.~S.; Barros,~K. Hierarchical modeling of molecular
  energies using a deep neural network. \emph{The Journal of chemical physics}
  \textbf{2018}, \emph{148}, 241715\relax
\mciteBstWouldAddEndPuncttrue
\mciteSetBstMidEndSepPunct{\mcitedefaultmidpunct}
{\mcitedefaultendpunct}{\mcitedefaultseppunct}\relax
\EndOfBibitem
\bibitem[Zubatyuk \latin{et~al.}(2019)Zubatyuk, Smith, Leszczynski, and
  Isayev]{zubatyuk2019accurate}
Zubatyuk,~R.; Smith,~J.~S.; Leszczynski,~J.; Isayev,~O. Accurate and
  transferable multitask prediction of chemical properties with an
  atoms-in-molecules neural network. \emph{Science advances} \textbf{2019},
  \emph{5}, eaav6490\relax
\mciteBstWouldAddEndPuncttrue
\mciteSetBstMidEndSepPunct{\mcitedefaultmidpunct}
{\mcitedefaultendpunct}{\mcitedefaultseppunct}\relax
\EndOfBibitem
\bibitem[Thomas \latin{et~al.}(2018)Thomas, Smidt, Kearnes, Yang, Li, Kohlhoff,
  and Riley]{thomas2018tensor}
Thomas,~N.; Smidt,~T.; Kearnes,~S.; Yang,~L.; Li,~L.; Kohlhoff,~K.; Riley,~P.
  Tensor field networks: Rotation-and translation-equivariant neural networks
  for 3d point clouds. \emph{arXiv preprint arXiv:1802.08219} \textbf{2018},
  \relax
\mciteBstWouldAddEndPunctfalse
\mciteSetBstMidEndSepPunct{\mcitedefaultmidpunct}
{}{\mcitedefaultseppunct}\relax
\EndOfBibitem
\bibitem[Anderson \latin{et~al.}(2019)Anderson, Hy, and
  Kondor]{anderson2019cormorant}
Anderson,~B.; Hy,~T.~S.; Kondor,~R. Cormorant: Covariant molecular neural
  networks. \emph{Advances in neural information processing systems}
  \textbf{2019}, \emph{32}\relax
\mciteBstWouldAddEndPuncttrue
\mciteSetBstMidEndSepPunct{\mcitedefaultmidpunct}
{\mcitedefaultendpunct}{\mcitedefaultseppunct}\relax
\EndOfBibitem
\bibitem[Fuchs \latin{et~al.}(2020)Fuchs, Worrall, Fischer, and
  Welling]{fuchs2020se}
Fuchs,~F.; Worrall,~D.; Fischer,~V.; Welling,~M. Se (3)-transformers: 3d
  roto-translation equivariant attention networks. \emph{Advances in Neural
  Information Processing Systems} \textbf{2020}, \emph{33}, 1970--1981\relax
\mciteBstWouldAddEndPuncttrue
\mciteSetBstMidEndSepPunct{\mcitedefaultmidpunct}
{\mcitedefaultendpunct}{\mcitedefaultseppunct}\relax
\EndOfBibitem
\bibitem[Brandstetter \latin{et~al.}(2022)Brandstetter, Hesselink, van~der Pol,
  Bekkers, and Welling]{brandstetter2022geometric}
Brandstetter,~J.; Hesselink,~R.; van~der Pol,~E.; Bekkers,~E.~J.; Welling,~M.
  Geometric and Physical Quantities improve E(3) Equivariant Message Passing.
  International Conference on Learning Representations. 2022\relax
\mciteBstWouldAddEndPuncttrue
\mciteSetBstMidEndSepPunct{\mcitedefaultmidpunct}
{\mcitedefaultendpunct}{\mcitedefaultseppunct}\relax
\EndOfBibitem
\bibitem[Jing \latin{et~al.}(2021)Jing, Eismann, Suriana, Townshend, and
  Dror]{jing2021learning}
Jing,~B.; Eismann,~S.; Suriana,~P.; Townshend,~R. J.~L.; Dror,~R. Learning from
  Protein Structure with Geometric Vector Perceptrons. International Conference
  on Learning Representations. 2021\relax
\mciteBstWouldAddEndPuncttrue
\mciteSetBstMidEndSepPunct{\mcitedefaultmidpunct}
{\mcitedefaultendpunct}{\mcitedefaultseppunct}\relax
\EndOfBibitem
\bibitem[Villar \latin{et~al.}(2021)Villar, Hogg, Storey-Fisher, Yao, and
  Blum-Smith]{villar2021scalars}
Villar,~S.; Hogg,~D.~W.; Storey-Fisher,~K.; Yao,~W.; Blum-Smith,~B. Scalars are
  universal: Equivariant machine learning, structured like classical physics.
  Advances in Neural Information Processing Systems. 2021\relax
\mciteBstWouldAddEndPuncttrue
\mciteSetBstMidEndSepPunct{\mcitedefaultmidpunct}
{\mcitedefaultendpunct}{\mcitedefaultseppunct}\relax
\EndOfBibitem
\bibitem[Gasteiger \latin{et~al.}(2021)Gasteiger, Becker, and
  G{\"u}nnemann]{gasteiger2021gemnet}
Gasteiger,~J.; Becker,~F.; G{\"u}nnemann,~S. Gemnet: Universal directional
  graph neural networks for molecules. \emph{Advances in Neural Information
  Processing Systems} \textbf{2021}, \emph{34}, 6790--6802\relax
\mciteBstWouldAddEndPuncttrue
\mciteSetBstMidEndSepPunct{\mcitedefaultmidpunct}
{\mcitedefaultendpunct}{\mcitedefaultseppunct}\relax
\EndOfBibitem
\bibitem[Batzner \latin{et~al.}(2022)Batzner, Musaelian, Sun, Geiger, Mailoa,
  Kornbluth, Molinari, Smidt, and Kozinsky]{batzner20223}
Batzner,~S.; Musaelian,~A.; Sun,~L.; Geiger,~M.; Mailoa,~J.~P.; Kornbluth,~M.;
  Molinari,~N.; Smidt,~T.~E.; Kozinsky,~B. E (3)-equivariant graph neural
  networks for data-efficient and accurate interatomic potentials. \emph{Nature
  communications} \textbf{2022}, \emph{13}, 1--11\relax
\mciteBstWouldAddEndPuncttrue
\mciteSetBstMidEndSepPunct{\mcitedefaultmidpunct}
{\mcitedefaultendpunct}{\mcitedefaultseppunct}\relax
\EndOfBibitem
\bibitem[Sch\"{u}tt \latin{et~al.}(2021)Sch\"{u}tt, Unke, and
  Gastegger]{schutt2021equivariant}
Sch\"{u}tt,~K.; Unke,~O.; Gastegger,~M. Equivariant message passing for the
  prediction of tensorial properties and molecular spectra. International
  Conference on Machine Learning. 2021; pp 9377--9388\relax
\mciteBstWouldAddEndPuncttrue
\mciteSetBstMidEndSepPunct{\mcitedefaultmidpunct}
{\mcitedefaultendpunct}{\mcitedefaultseppunct}\relax
\EndOfBibitem
\bibitem[Th{\"o}lke and De~Fabritiis(2022)Th{\"o}lke, and
  De~Fabritiis]{tholke2022torchmd}
Th{\"o}lke,~P.; De~Fabritiis,~G. TorchMD-NET: Equivariant Transformers for
  Neural Network based Molecular Potentials. \emph{arXiv preprint
  arXiv:2202.02541} \textbf{2022}, \relax
\mciteBstWouldAddEndPunctfalse
\mciteSetBstMidEndSepPunct{\mcitedefaultmidpunct}
{}{\mcitedefaultseppunct}\relax
\EndOfBibitem
\bibitem[Vaswani \latin{et~al.}(2017)Vaswani, Shazeer, Parmar, Uszkoreit,
  Jones, Gomez, Kaiser, and Polosukhin]{vaswani2017attention}
Vaswani,~A.; Shazeer,~N.; Parmar,~N.; Uszkoreit,~J.; Jones,~L.; Gomez,~A.~N.;
  Kaiser,~{\L}.; Polosukhin,~I. Attention is all you need. \emph{Advances in
  neural information processing systems} \textbf{2017}, \emph{30}\relax
\mciteBstWouldAddEndPuncttrue
\mciteSetBstMidEndSepPunct{\mcitedefaultmidpunct}
{\mcitedefaultendpunct}{\mcitedefaultseppunct}\relax
\EndOfBibitem
\bibitem[Hadsell \latin{et~al.}(2006)Hadsell, Chopra, and
  LeCun]{hadsell2006dimensionality}
Hadsell,~R.; Chopra,~S.; LeCun,~Y. Dimensionality reduction by learning an
  invariant mapping. 2006 IEEE Computer Society Conference on Computer Vision
  and Pattern Recognition (CVPR'06). 2006; pp 1735--1742\relax
\mciteBstWouldAddEndPuncttrue
\mciteSetBstMidEndSepPunct{\mcitedefaultmidpunct}
{\mcitedefaultendpunct}{\mcitedefaultseppunct}\relax
\EndOfBibitem
\bibitem[Chen \latin{et~al.}(2020)Chen, Kornblith, Norouzi, and
  Hinton]{chen2020simple}
Chen,~T.; Kornblith,~S.; Norouzi,~M.; Hinton,~G. A simple framework for
  contrastive learning of visual representations. International conference on
  machine learning. 2020; pp 1597--1607\relax
\mciteBstWouldAddEndPuncttrue
\mciteSetBstMidEndSepPunct{\mcitedefaultmidpunct}
{\mcitedefaultendpunct}{\mcitedefaultseppunct}\relax
\EndOfBibitem
\bibitem[Hu* \latin{et~al.}(2020)Hu*, Liu*, Gomes, Zitnik, Liang, Pande, and
  Leskovec]{Hu2020Strategies}
Hu*,~W.; Liu*,~B.; Gomes,~J.; Zitnik,~M.; Liang,~P.; Pande,~V.; Leskovec,~J.
  Strategies for Pre-training Graph Neural Networks. International Conference
  on Learning Representations. 2020\relax
\mciteBstWouldAddEndPuncttrue
\mciteSetBstMidEndSepPunct{\mcitedefaultmidpunct}
{\mcitedefaultendpunct}{\mcitedefaultseppunct}\relax
\EndOfBibitem
\bibitem[Rong \latin{et~al.}(2020)Rong, Bian, Xu, Xie, Wei, Huang, and
  Huang]{rong2020self}
Rong,~Y.; Bian,~Y.; Xu,~T.; Xie,~W.; Wei,~Y.; Huang,~W.; Huang,~J.
  Self-supervised graph transformer on large-scale molecular data.
  \emph{Advances in Neural Information Processing Systems} \textbf{2020},
  \emph{33}, 12559--12571\relax
\mciteBstWouldAddEndPuncttrue
\mciteSetBstMidEndSepPunct{\mcitedefaultmidpunct}
{\mcitedefaultendpunct}{\mcitedefaultseppunct}\relax
\EndOfBibitem
\bibitem[Zhang \latin{et~al.}(2021)Zhang, Liu, Wang, Lu, and
  Lee]{zhang2021motif}
Zhang,~Z.; Liu,~Q.; Wang,~H.; Lu,~C.; Lee,~C.-K. Motif-based graph
  self-supervised learning for molecular property prediction. \emph{Advances in
  Neural Information Processing Systems} \textbf{2021}, \emph{34},
  15870--15882\relax
\mciteBstWouldAddEndPuncttrue
\mciteSetBstMidEndSepPunct{\mcitedefaultmidpunct}
{\mcitedefaultendpunct}{\mcitedefaultseppunct}\relax
\EndOfBibitem
\bibitem[Fang \latin{et~al.}(2022)Fang, Liu, Lei, He, Zhang, Zhou, Wang, Wu,
  and Wang]{fang2022geometry}
Fang,~X.; Liu,~L.; Lei,~J.; He,~D.; Zhang,~S.; Zhou,~J.; Wang,~F.; Wu,~H.;
  Wang,~H. Geometry-enhanced molecular representation learning for property
  prediction. \emph{Nature Machine Intelligence} \textbf{2022}, \emph{4},
  127--134\relax
\mciteBstWouldAddEndPuncttrue
\mciteSetBstMidEndSepPunct{\mcitedefaultmidpunct}
{\mcitedefaultendpunct}{\mcitedefaultseppunct}\relax
\EndOfBibitem
\bibitem[Liu \latin{et~al.}(2022)Liu, Jin, Pan, Zhou, Zheng, Xia, and
  Yu]{liu2022graph}
Liu,~Y.; Jin,~M.; Pan,~S.; Zhou,~C.; Zheng,~Y.; Xia,~F.; Yu,~P. Graph
  self-supervised learning: A survey. \emph{IEEE Transactions on Knowledge and
  Data Engineering} \textbf{2022}, \relax
\mciteBstWouldAddEndPunctfalse
\mciteSetBstMidEndSepPunct{\mcitedefaultmidpunct}
{}{\mcitedefaultseppunct}\relax
\EndOfBibitem
\bibitem[Krishnan \latin{et~al.}(2022)Krishnan, Rajpurkar, and
  Topol]{krishnan2022self}
Krishnan,~R.; Rajpurkar,~P.; Topol,~E.~J. Self-supervised learning in medicine
  and healthcare. \emph{Nature Biomedical Engineering} \textbf{2022},
  1--7\relax
\mciteBstWouldAddEndPuncttrue
\mciteSetBstMidEndSepPunct{\mcitedefaultmidpunct}
{\mcitedefaultendpunct}{\mcitedefaultseppunct}\relax
\EndOfBibitem
\bibitem[Magar \latin{et~al.}(2022)Magar, Wang, and
  Barati~Farimani]{magar2022crystal}
Magar,~R.; Wang,~Y.; Barati~Farimani,~A. Crystal twins: self-supervised
  learning for crystalline material property prediction. \emph{npj
  Computational Materials} \textbf{2022}, \emph{8}, 1--8\relax
\mciteBstWouldAddEndPuncttrue
\mciteSetBstMidEndSepPunct{\mcitedefaultmidpunct}
{\mcitedefaultendpunct}{\mcitedefaultseppunct}\relax
\EndOfBibitem
\bibitem[Cao \latin{et~al.}(2022)Cao, Magar, Wang, and
  Farimani]{cao2022moformer}
Cao,~Z.; Magar,~R.; Wang,~Y.; Farimani,~A.~B. MOFormer: Self-Supervised
  Transformer model for Metal-Organic Framework Property Prediction.
  \emph{arXiv preprint arXiv:2210.14188} \textbf{2022}, \relax
\mciteBstWouldAddEndPunctfalse
\mciteSetBstMidEndSepPunct{\mcitedefaultmidpunct}
{}{\mcitedefaultseppunct}\relax
\EndOfBibitem
\bibitem[Wang \latin{et~al.}(2022)Wang, Wang, Cao, and
  Barati~Farimani]{wang2022molecular}
Wang,~Y.; Wang,~J.; Cao,~Z.; Barati~Farimani,~A. Molecular contrastive learning
  of representations via graph neural networks. \emph{Nature Machine
  Intelligence} \textbf{2022}, \emph{4}, 279--287\relax
\mciteBstWouldAddEndPuncttrue
\mciteSetBstMidEndSepPunct{\mcitedefaultmidpunct}
{\mcitedefaultendpunct}{\mcitedefaultseppunct}\relax
\EndOfBibitem
\bibitem[Zhang \latin{et~al.}(2020)Zhang, Hu, Subramonian, and
  Sun]{zhang2020motif}
Zhang,~S.; Hu,~Z.; Subramonian,~A.; Sun,~Y. Motif-driven contrastive learning
  of graph representations. \emph{arXiv preprint arXiv:2012.12533}
  \textbf{2020}, \relax
\mciteBstWouldAddEndPunctfalse
\mciteSetBstMidEndSepPunct{\mcitedefaultmidpunct}
{}{\mcitedefaultseppunct}\relax
\EndOfBibitem
\bibitem[Wang \latin{et~al.}(2022)Wang, Magar, Liang, and
  Barati~Farimani]{wang2022improving}
Wang,~Y.; Magar,~R.; Liang,~C.; Barati~Farimani,~A. Improving Molecular
  Contrastive Learning via Faulty Negative Mitigation and Decomposed Fragment
  Contrast. \emph{Journal of Chemical Information and Modeling} \textbf{2022},
  \relax
\mciteBstWouldAddEndPunctfalse
\mciteSetBstMidEndSepPunct{\mcitedefaultmidpunct}
{}{\mcitedefaultseppunct}\relax
\EndOfBibitem
\bibitem[Liu \latin{et~al.}(2022)Liu, Wang, Liu, Lasenby, Guo, and
  Tang]{liu2022pretraining}
Liu,~S.; Wang,~H.; Liu,~W.; Lasenby,~J.; Guo,~H.; Tang,~J. Pre-training
  Molecular Graph Representation with 3D Geometry. International Conference on
  Learning Representations. 2022\relax
\mciteBstWouldAddEndPuncttrue
\mciteSetBstMidEndSepPunct{\mcitedefaultmidpunct}
{\mcitedefaultendpunct}{\mcitedefaultseppunct}\relax
\EndOfBibitem
\bibitem[St{\"a}rk \latin{et~al.}(2022)St{\"a}rk, Beaini, Corso, Tossou,
  Dallago, G{\"u}nnemann, and Li{\`o}]{stark20223d}
St{\"a}rk,~H.; Beaini,~D.; Corso,~G.; Tossou,~P.; Dallago,~C.;
  G{\"u}nnemann,~S.; Li{\`o},~P. 3d infomax improves gnns for molecular
  property prediction. International Conference on Machine Learning. 2022; pp
  20479--20502\relax
\mciteBstWouldAddEndPuncttrue
\mciteSetBstMidEndSepPunct{\mcitedefaultmidpunct}
{\mcitedefaultendpunct}{\mcitedefaultseppunct}\relax
\EndOfBibitem
\bibitem[Zaidi \latin{et~al.}(2022)Zaidi, Schaarschmidt, Martens, Kim, Teh,
  Sanchez-Gonzalez, Battaglia, Pascanu, and Godwin]{zaidi2022pre}
Zaidi,~S.; Schaarschmidt,~M.; Martens,~J.; Kim,~H.; Teh,~Y.~W.;
  Sanchez-Gonzalez,~A.; Battaglia,~P.; Pascanu,~R.; Godwin,~J. Pre-training via
  Denoising for Molecular Property Prediction. \emph{arXiv preprint
  arXiv:2206.00133} \textbf{2022}, \relax
\mciteBstWouldAddEndPunctfalse
\mciteSetBstMidEndSepPunct{\mcitedefaultmidpunct}
{}{\mcitedefaultseppunct}\relax
\EndOfBibitem
\bibitem[Liu \latin{et~al.}(2022)Liu, Guo, and Tang]{liu2022molecular}
Liu,~S.; Guo,~H.; Tang,~J. Molecular geometry pretraining with se (3)-invariant
  denoising distance matching. \emph{arXiv preprint arXiv:2206.13602}
  \textbf{2022}, \relax
\mciteBstWouldAddEndPunctfalse
\mciteSetBstMidEndSepPunct{\mcitedefaultmidpunct}
{}{\mcitedefaultseppunct}\relax
\EndOfBibitem
\bibitem[Zhou \latin{et~al.}(2023)Zhou, Gao, Ding, Zheng, Xu, Wei, Zhang, and
  Ke]{zhou2023unimol}
Zhou,~G.; Gao,~Z.; Ding,~Q.; Zheng,~H.; Xu,~H.; Wei,~Z.; Zhang,~L.; Ke,~G.
  Uni-Mol: A Universal 3D Molecular Representation Learning Framework. The
  Eleventh International Conference on Learning Representations. 2023\relax
\mciteBstWouldAddEndPuncttrue
\mciteSetBstMidEndSepPunct{\mcitedefaultmidpunct}
{\mcitedefaultendpunct}{\mcitedefaultseppunct}\relax
\EndOfBibitem
\bibitem[Ruddigkeit \latin{et~al.}(2012)Ruddigkeit, Van~Deursen, Blum, and
  Reymond]{ruddigkeit2012enumeration}
Ruddigkeit,~L.; Van~Deursen,~R.; Blum,~L.~C.; Reymond,~J.-L. Enumeration of 166
  billion organic small molecules in the chemical universe database GDB-17.
  \emph{Journal of chemical information and modeling} \textbf{2012}, \emph{52},
  2864--2875\relax
\mciteBstWouldAddEndPuncttrue
\mciteSetBstMidEndSepPunct{\mcitedefaultmidpunct}
{\mcitedefaultendpunct}{\mcitedefaultseppunct}\relax
\EndOfBibitem
\bibitem[Satorras \latin{et~al.}(2021)Satorras, Hoogeboom, and
  Welling]{satorras2021n}
Satorras,~V.~G.; Hoogeboom,~E.; Welling,~M. E (n) equivariant graph neural
  networks. International conference on machine learning. 2021; pp
  9323--9332\relax
\mciteBstWouldAddEndPuncttrue
\mciteSetBstMidEndSepPunct{\mcitedefaultmidpunct}
{\mcitedefaultendpunct}{\mcitedefaultseppunct}\relax
\EndOfBibitem
\bibitem[Xu \latin{et~al.}(2019)Xu, Hu, Leskovec, and Jegelka]{xu2018how}
Xu,~K.; Hu,~W.; Leskovec,~J.; Jegelka,~S. How Powerful are Graph Neural
  Networks? International Conference on Learning Representations. 2019\relax
\mciteBstWouldAddEndPuncttrue
\mciteSetBstMidEndSepPunct{\mcitedefaultmidpunct}
{\mcitedefaultendpunct}{\mcitedefaultseppunct}\relax
\EndOfBibitem
\bibitem[Xie \latin{et~al.}(2022)Xie, Fu, Ganea, Barzilay, and
  Jaakkola]{xie2022crystal}
Xie,~T.; Fu,~X.; Ganea,~O.-E.; Barzilay,~R.; Jaakkola,~T.~S. Crystal Diffusion
  Variational Autoencoder for Periodic Material Generation. International
  Conference on Learning Representations. 2022\relax
\mciteBstWouldAddEndPuncttrue
\mciteSetBstMidEndSepPunct{\mcitedefaultmidpunct}
{\mcitedefaultendpunct}{\mcitedefaultseppunct}\relax
\EndOfBibitem
\bibitem[Arts \latin{et~al.}(2023)Arts, Satorras, Huang, Zuegner, Federici,
  Clementi, No{\'e}, Pinsler, and Berg]{arts2023two}
Arts,~M.; Satorras,~V.~G.; Huang,~C.-W.; Zuegner,~D.; Federici,~M.;
  Clementi,~C.; No{\'e},~F.; Pinsler,~R.; Berg,~R. v.~d. Two for One: Diffusion
  Models and Force Fields for Coarse-Grained Molecular Dynamics. \emph{arXiv
  preprint arXiv:2302.00600} \textbf{2023}, \relax
\mciteBstWouldAddEndPunctfalse
\mciteSetBstMidEndSepPunct{\mcitedefaultmidpunct}
{}{\mcitedefaultseppunct}\relax
\EndOfBibitem
\bibitem[Smith \latin{et~al.}(2020)Smith, Zubatyuk, Nebgen, Lubbers, Barros,
  Roitberg, Isayev, and Tretiak]{smith2020ani}
Smith,~J.~S.; Zubatyuk,~R.; Nebgen,~B.; Lubbers,~N.; Barros,~K.;
  Roitberg,~A.~E.; Isayev,~O.; Tretiak,~S. The ANI-1ccx and ANI-1x data sets,
  coupled-cluster and density functional theory properties for molecules.
  \emph{Scientific data} \textbf{2020}, \emph{7}, 1--10\relax
\mciteBstWouldAddEndPuncttrue
\mciteSetBstMidEndSepPunct{\mcitedefaultmidpunct}
{\mcitedefaultendpunct}{\mcitedefaultseppunct}\relax
\EndOfBibitem
\bibitem[Sch{\"u}tt \latin{et~al.}(2017)Sch{\"u}tt, Kindermans, Sauceda~Felix,
  Chmiela, Tkatchenko, and M{\"u}ller]{schutt2017schnet}
Sch{\"u}tt,~K.; Kindermans,~P.-J.; Sauceda~Felix,~H.~E.; Chmiela,~S.;
  Tkatchenko,~A.; M{\"u}ller,~K.-R. Schnet: A continuous-filter convolutional
  neural network for modeling quantum interactions. \emph{Advances in neural
  information processing systems} \textbf{2017}, \emph{30}\relax
\mciteBstWouldAddEndPuncttrue
\mciteSetBstMidEndSepPunct{\mcitedefaultmidpunct}
{\mcitedefaultendpunct}{\mcitedefaultseppunct}\relax
\EndOfBibitem
\bibitem[Eastman \latin{et~al.}(2023)Eastman, Behara, Dotson, Galvelis, Herr,
  Horton, Mao, Chodera, Pritchard, Wang, \latin{et~al.}
  others]{eastman2023spice}
Eastman,~P.; Behara,~P.~K.; Dotson,~D.~L.; Galvelis,~R.; Herr,~J.~E.;
  Horton,~J.~T.; Mao,~Y.; Chodera,~J.~D.; Pritchard,~B.~P.; Wang,~Y.,
  \latin{et~al.}  SPICE, A Dataset of Drug-like Molecules and Peptides for
  Training Machine Learning Potentials. \emph{Scientific Data} \textbf{2023},
  \emph{10}, 1--11\relax
\mciteBstWouldAddEndPuncttrue
\mciteSetBstMidEndSepPunct{\mcitedefaultmidpunct}
{\mcitedefaultendpunct}{\mcitedefaultseppunct}\relax
\EndOfBibitem
\bibitem[Chmiela \latin{et~al.}(2023)Chmiela, Vassilev-Galindo, Unke, Kabylda,
  Sauceda, Tkatchenko, and Müller]{chmiela2023accurate}
Chmiela,~S.; Vassilev-Galindo,~V.; Unke,~O.~T.; Kabylda,~A.; Sauceda,~H.~E.;
  Tkatchenko,~A.; Müller,~K.-R. Accurate global machine learning force fields
  for molecules with hundreds of atoms. \emph{Science Advances} \textbf{2023},
  \emph{9}, eadf0873\relax
\mciteBstWouldAddEndPuncttrue
\mciteSetBstMidEndSepPunct{\mcitedefaultmidpunct}
{\mcitedefaultendpunct}{\mcitedefaultseppunct}\relax
\EndOfBibitem
\bibitem[Loshchilov and Hutter(2017)Loshchilov, and
  Hutter]{loshchilov2017decoupled}
Loshchilov,~I.; Hutter,~F. Decoupled weight decay regularization. \emph{arXiv
  preprint arXiv:1711.05101} \textbf{2017}, \relax
\mciteBstWouldAddEndPunctfalse
\mciteSetBstMidEndSepPunct{\mcitedefaultmidpunct}
{}{\mcitedefaultseppunct}\relax
\EndOfBibitem
\bibitem[Loshchilov and Hutter(2017)Loshchilov, and Hutter]{loshchilov2017sgdr}
Loshchilov,~I.; Hutter,~F. {SGDR}: Stochastic Gradient Descent with Warm
  Restarts. International Conference on Learning Representations. 2017\relax
\mciteBstWouldAddEndPuncttrue
\mciteSetBstMidEndSepPunct{\mcitedefaultmidpunct}
{\mcitedefaultendpunct}{\mcitedefaultseppunct}\relax
\EndOfBibitem
\end{mcitethebibliography}

\providecommand{\latin}[1]{#1}
\makeatletter
\providecommand{\doi}
  {\begingroup\let\do\@makeother\dospecials
  \catcode`\{=1 \catcode`\}=2 \doi@aux}
\providecommand{\doi@aux}[1]{\endgroup\texttt{#1}}
\makeatother
\providecommand*\mcitethebibliography{\thebibliography}
\csname @ifundefined\endcsname{endmcitethebibliography}
  {\let\endmcitethebibliography\endthebibliography}{}



\end{document}